\documentclass[preprint,3p,twocolumn, hidelinks]{elsarticle}

\usepackage{lineno,hyperref}
\usepackage{xcolor}
\modulolinenumbers[5]

\usepackage{graphicx} 
\usepackage{amsmath}
\usepackage{amsthm}
\usepackage{mathtools}
\usepackage{amssymb}
\usepackage{bm}
\usepackage{fontawesome}
\usepackage[toc,page]{appendix}

\usepackage{tikz}
\usetikzlibrary{shapes.geometric, arrows}
\usetikzlibrary{decorations.pathreplacing}
\usetikzlibrary{decorations.text}

\usepackage{fix-cm}    
\makeatletter
\newcommand\HUGE{\@setfontsize\Huge{45}{45}}
\makeatother

\newcommand{\x}{\textbf{x}}

\newcommand{\y}{\textbf{y}}

\newcommand{\w}{\textbf{w}}

\newcommand{\X}{\textbf{X}}

\newtheorem{thm}{Theorem}[section]

\newtheorem{prop}[thm]{Proposition}

\newtheorem{defn}[thm]{Definition}
\newtheorem{assumption}[thm]{Assumption}
\newtheorem{ex}[thm]{Example}
\newtheorem{remark}[thm]{Remark}

\def\eqnref#1{(\ref{#1})}

\def\exref#1{Example \ref{#1}}

\def\figref#1{Figure \ref{#1}}
\def\secref#1{Section \ref{#1}}

\def\appref#1{Appendix \ref{#1}}

\journal{Reliability Engineering \& System Safety}

\makeatletter
\def\ps@pprintTitle{%
 \let\@oddhead\@empty
 \let\@evenhead\@empty
 \def\@oddfoot{}%
 \let\@evenfoot\@oddfoot}
\makeatother
\newcommand\blfootnote[1]{%
  \begingroup
  \renewcommand\thefootnote{}\footnote{#1}%
  \addtocounter{footnote}{-1}%
  \endgroup
}

\begin{document}

\begin{frontmatter}

\title{Optimal sequential decision making
with probabilistic digital twins}

\author[dnvgl,uio]{Christian Agrell}
\author[uio]{Kristina Rognlien Dahl}
\author[dnvgl]{Andreas Hafver}

\address[dnvgl]{Group Research and Development, DNV, Norway}
\address[uio]{Department of Mathematics, University of Oslo, Norway}

\begin{abstract}
Digital twins are emerging in many industries, typically 
consisting of simulation models and data associated with a specific physical system. 
One of the main reasons for developing a digital twin, is to enable the simulation of possible consequences 
of a given action, without the need to interfere with the physical system itself. 
Physical systems of interest, and the environments they operate in, do not always behave deterministically. Moreover, information about the system and its environment is typically incomplete or imperfect. Probabilistic representations of systems and environments may therefore be called for, especially to support decisions in application areas where actions may have severe consequences. A \emph{probabilistic digital twin} is a digital twin,  with the added capability of proper treatment of uncertainties associated with the consequences, enabling better decision support and management of risks.

In this paper we introduce the probabilistic digital twin (PDT).
We will start by discussing how epistemic uncertainty can be treated using measure theory, by modelling epistemic information via $\sigma$-algebras. Based on this, we give a formal definition of how epistemic uncertainty can be updated in a PDT.
We then study the problem of optimal sequential decision making. That is, we consider the case where the outcome of each decision may inform the next.
Within the PDT framework, we formulate this optimization problem. We discuss how this problem may be solved (at least in theory) via the maximum principle method or the dynamic programming principle. However, due to the curse of dimensionality, these methods are often not tractable in practice. To mend this, we propose a generic 
approximate solution using deep reinforcement learning together with neural networks defined on sets. 
We illustrate the method on a practical problem, considering optimal information gathering for the estimation of a failure probability. 
\end{abstract}


\end{frontmatter}

\blfootnote{\textbf{pre-print:} This is a pre-print version of this article}


\section{Introduction}

The use of digital twins has emerged as one of the major technology trends the last couple of years.
In essence, a digital twin (DT) is a digital representation of some physical system, including data from observations of the physical system, which can be used  
 to 
perform forecasts, evaluate the consequences of potential actions, simulate possible future scenarios, 
and in general inform decision making without requiring interference with the physical system. 
From a theoretical perspective, a digital twin may be regraded to consist of the following two components: 
\begin{itemize}
    \item A set of \emph{assumptions} regarding the physical system (e.g. about the behaviour or relationships among system components and between the system and its environment), often given in the form of a physics-based 
    numerical simulation model.
    \item A set of \emph{information}, usually in the form of a set of observations, 
    or records of the relevant actions taken within the system. 
\end{itemize}

In some cases, a digital twin may be desired for a system which attributes and behaviours are not deterministic, but stochastic. For example, the degradation and failure of physical structures or machinery is typically described as stochastic processes. A systems performance may be impacted by weather or financial conditions, which also may be most appropriately modelled as stochastic. Sometimes the functioning of the system itself is stochastic, such as supply chain or production chains involving stochastic variation in demand and performance of various system components.  

Even for systems or phenomena that are deterministic in principle, a model will never give a perfect rendering of reality. There will typically be uncertainty about the model’s structure and parameters (i.e. epistemic uncertainty), and if consequences of actions can be critical, such uncertainties need to be captured and handled appropriately by the digital twin. In general, the system of interest will have both stochastic elements (aleatory uncertainty) and epistemic uncertainty. 

If we want to apply digital twins to inform decisions in systems where the analysis of uncertainty and risk is important, 
certain properties are required:
\begin{enumerate}
    \item The digital twin must capture uncertainties: \textit{This could be done by using a probabilistic representation for uncertain system attributes.}
    \item It should be possible to update the digital twin as new information becomes available: \textit{This could be from new 
    evidence in the form of data, or underlying assumptions about the system that have changed.}
    \item For the digital twin to be informative in decision making, it should be possible to query the model sufficiently fast: 
    \textit{This could mean making use of surrogate models or emulators, which introduces additional uncertainties. }
\end{enumerate}

These properties are paraphrased from Hafver et al. \cite{Hafver:2018:PDT}, which provides a detailed discussion on the use 
of digital twins for on-line risk assessment. 
In this paper we propose a mathematical framework for defining digital twins that comply with these properties.
As Hafver et al. \cite{Hafver:2018:PDT}, we will refer to these as \emph{probabilistic digital twins} (PDTs), 
and we will build on the Bayesian probabilistic framework which is a natural choice
to satisfy (1)-(2).

A numerical model of a complex physical system can often be computationally expensive, 
for instance if it involves numerical solution of nontrivial partial differential equations. 
In a probabilistic setting this is prohibitive, as a large number of evaluations (e.g. PDE solves) 
is needed for tasks involving uncertainty propagation, such as prediction and inference.
Applications towards real-time decision making also sets natural restrictions with respect to 
the runtime of such queries. 
This is why property (3) is important, and why probabilistic models of complex physical 
phenomena often involve the use of approximate alternatives, usually obtained by "fitting" a 
computationally cheap model to the output of a few expensive model runs. 
These computationally cheap approximations are often referred to as 
\textit{response surface models}, \textit{surrogate models} or \textit{emulators} in the literature. 

Introducing this kind of approximation for computational efficiency also means that we introduce additional epistemic uncertainty into our modelling framework. 
By \emph{epistemic} uncertainty we mean, in short, any form of uncertainty that can be reduced by gathering more information (to be discussed further later on). In our context, uncertainty may in principle be reduced by running the expensive numerical modes instead of the cheaper approximations.

Many interesting sequential decision making problems arise from the property that our knowledge about the 
system we operate changes as we learn about the outcomes. That is, each decision may affect the epistemic uncertainty 
which the next decision will be based upon. 
We are motivated by this type of scenario, in combination with 
the challenge of finding robust decisions in safety-critical systems, where a decision 
should be robust with respect to what we do not know, i.e. with respect to epistemic uncertainty. 
Although we will not restrict the framework presented in this paper to any specific type of sequential decision making objectives, 
we will mainly focus on problems related to \emph{optimal information gathering}. 
That is, where the decisions we consider are related to acquiring information
(e.g., by running an experiment) in order to reduce the epistemic uncertainty with respect to some 
specified objective (e.g., estimating some quantity of interest). 

A very relevant example of such a task, is the problem of optimal experimental design for structural reliability analysis. This involves 
deciding which experiments to run in order to build a surrogate model that can be used to estimate 
a failure probability with sufficient level of confidence. 
This is a problem that has received considerable attention 
(see e.g. \citep{Agrell:2020:DOE_SRA, Bect:2012:Sequential_design, Echard:2011:AKMCS, Bichon:2008:EGRA, Sun:2017:LIF, Jian:2017:two_acc, Perrin:2016:AL_multiple_fm}).
These methods all make use of a myopic (one-step lookahead) criterion to determine the "optimal" experiment, 
as a multi-step or full dynamic programming formulation of the optimization problem becomes numerically infeasible. In 
Agrell and Dahl \cite{Agrell:2020:DOE_SRA}, they consider the case where there are different types of experiments to choose from. 
Here, the myopic (one-step lookahead) assumption can still be justified, but if the different types of experiments are associated with different costs,
then it can be difficult to apply in practice (e.g., if a feasible solution requires expensive experiments with delayed reward).  

We will review the mathematical framework of sequential decision making, and connect this to the definition of a PDT. Traditionally, there are two main solution strategies for solving discrete time sequential decision making problems: Maximum principles, and dynamic programming. We review these two solution methods, and conclude that the PDT framework is well suited for a dynamic programming approach. However, dynamic programming suffers from the curse of dimensionality, i.e. possible sequences of decisions and state realizations grow exponentially with the size of the state space. Hence, we are typically not able to solve a PDT sequential decision making problem in practice directly via dynamic programming. 

As a generic solution to 
the problem of optimal sequential decision making we instead propose an alternative based on reinforcement learning.
This means that when we consider the problem of finding an optimal decision policy, instead of 
truncating the theoretical optimal solution (from the Bellman equation) by e.g., looking only one step ahead, we try to approximate 
the optimal policy. This approximation can be done by using e.g. a neural network. Here we will frame the sequential decision making setup as a Markov 
decision process (MDP), in general as a \emph{partially observed} MDP (POMDP), where a \emph{state} 
is represented by the information available at any given time. This kind of state specification is often referred to 
as the \emph{information state-space}. As a generic approach to deep reinforcement learning using PDTs, we propose an approach using neural networks that operate on the information state-space directly.  

\bigskip

\textbf{Main contributions.} In this paper we will:

\begin{enumerate}
    \item[$(i)$] Propose a mathematical framework for modelling epistemic uncertainty based on measure theory, and define epistemic conditioning. 
    \item[$(ii)$] Present a mathematical definition of the probabilistic digital twin (PDT). This is a mathematical framework for modelling physical systems with aleatory and epistemic uncertainty. 
    \item[$(iii)$] Introduce the problem of sequential decision making in the PDT, and illustrate how this problem can be solved (at least in theory) via maximum principle methods or the dynamic programming principle. 
    \item[$(iv)$] Discuss the curse of dimensionality for these solution methods, and illustrate how the sequential decision making problem in the PDT can be viewed as a partially observable Markov decision process.
    \item[$(v)$] Explain how reinforcement learning (RL) can be applied to find approximate optimal strategies for sequential decision making in the PDT, and propose a generic approach using a deep sets architecture that enables RL directly on the information state-space. We end with a numerical example to illustrate this approach.
\end{enumerate} 

The paper is structured as follows: In Section \ref{sec:measure_epistemic} we introduce epistemic uncertainty and suggest modeling this via $\sigma$-algebras. We also define epistemic conditioning. In Section \ref{sec:PDT}, we present the mathematical framework, as well as a formal definition, of a probabilistic digital twin (PDT), and discuss how such PDTs are used in practice. 

Then, in Section \ref{sec:seq_dec_making}, we introduce the problem of stochastic sequential decision making. We discuss the traditional solution approaches, in particular dynamic programming
which is theoretically a suitable approach for decision problems that can be modelled using a PDT. 
However, due to the curse of dimensionality, using the dynamic programming directly is typically not tractable.
We therefore turn to reinforcement learning using function approximation as a practical alternative. 
In Section \ref{sec: deep_RL}, we show how an approximate optimal strategy can be achieved using deep reinforcement learning, and
we illustrate the approach with a numerical example. Finally, in Section \ref{sec: conclusion} we conclude and sketch some future works in this direction.

\section{A measure-theoretic treatment of epistemic uncertainty}
\label{sec:measure_epistemic}

In this section, we review the concepts of epistemic and aleatory uncertainty, and introduce a measure-theoretic framework for modelling epistemic uncertainty. We will also define epistemic conditioning.

\subsection{Motivation}
In uncertainty quantification (UQ), it is common to consider two different kinds of uncertainty: Aleatory (stochastic) and epistemic (knowledge-based) uncertainty. 
We say that uncertainty is epistemic if we foresee the possibility of reducing it through gathering more or better information. 
For instance, uncertainty related to a parameter that has a fixed but unknown value is considered epistemic. 
Aleatory uncertainty, on the other hand, is the uncertainty which cannot (in the modellers perspective) be affected by gathering information alone. 
Note that the characterization of aleatory and epistemic uncertainty has to depend on the modelling context. 
For instance, the result of a coin flip may be viewed as epistemic, if we imagine a physics-based model 
that could predict the outcome exactly (given all initial conditions etc.). 
However, under most circumstances it is most natural to view a coin flip as aleatory, or that it contains both aleatory and epistemic uncertainty 
(e.g. if the bias of the coin us unknown).
Der Kiureghian et al. \cite{AorE} provides a detailed discussion of the differences between aleatory and epistemic uncertainty.

In this paper, we have two main reasons for distinguishing between epistemic and aleatory uncertainty. First, we would like to make decisions that are robust with respect to epistemic uncertainty. Secondly, we are interested in studying the effect of gathering information. Modelling epistemic uncertainty is a natural way of doing this. 

In the UQ literature, aleatory uncertainty is typically modelled via probability theory. 
However, epistemic uncertainty is represented in many different ways. 
For instance, Helton \cite{Sallaberry} considers four different ways of modelling epistemic uncertainty: Interval analysis, possibility theory, evidence theory (Dempster–Shafer theory) and probability theory.

In this paper we take a measure-theoretic approach. This provides a framework that is relatively flexible with respect to the types of assumptions that underlay the epistemic uncertainty. 
As a motivating example, consider the following typical setup used in statistics: 
\medskip
\begin{ex}(A parametric model)
\label{ex:X_theta}\\ 
    \noindent Let $\X = (Y, \theta)$ where $Y$ is a random variable representing some stochastic phenomenon, and assume $Y$ is modelled using a given probability distribution, $P(Y | \theta)$, that depends on a parameter $\theta$ (e.g. $Y \sim \mathcal{N}(\mu, \sigma)$ with $\theta = (\mu, \sigma)$). 
    Assume that we do not know the value of $\theta$, and we therefore consider $\theta$ as a (purely) epistemic parameter. For some fixed value of $\theta$, the random variable $Y$ is (purely) aleatory, 
    but in general, as the true value of $\theta$ is not known, $Y$ is associated with both epistemic and aleatory uncertainty. 
\end{ex}
\medskip
The model $\X$ in \exref{ex:X_theta} can be decoupled into an aleatory component $Y|\theta$ and an epistemic component $\theta$.
Any property of the aleatory uncertainty in $\X$ is determined by $P(Y | \theta)$, and is therefore a function of $\theta$. 
For instance, the probability $P(Y \in A | \theta)$ and the expectation $E[f(Y) | \theta]$, are both \emph{functions of} $\theta$. 
There are different ways in which we can choose to address the epistemic uncertainty in $\theta$. 
We could consider intervals, for instance the minimum and maximum of $P(Y \in A | \theta)$ over any plausible value of $\theta$, 
or assign probabilities, or some other measure of belief, to the possible values $\theta$ may take. 
However, in order for this to be well-defined mathematically, we need to put some requirements on $A$, $f$ and $\theta$. 
By using probability theory to represent the aleatory uncertainty, we implicitly assume that the set $A$ and function $f$ are \emph{measurable}, 
and we will assume that the same holds for $\theta$. We will describe in detail what is meant by measurable in \secref{sec:probspace} below. Essentially, this is just a necessity for defining properties such as distance, volume or probability in the space where $\theta$ resides.  

In this paper we will rely on probability theory for handling both aleatory and epistemic uncertainty. 
This means that, along with the measurability requirement on $\theta$, we have the familiar setup for Bayesian inference:
\medskip
\begin{ex}
\label{ex:X_theta_2} (A parametric model -- Inference and prediction)
    If $\theta$ from \exref{ex:X_theta} is a random variable with distribution $P(\theta)$, then $\X = (Y, \theta)$ denotes a complete probabilistic model (capturing both aleatory and epistemic uncertainty). 
    $\X$ is a random variable with distribution 
    \begin{equation*}
        P(\X) = P(Y | \theta)P(\theta).
    \end{equation*}
    Let $I$ be some piece of information from which Bayesian inference is possible, 
    i.e. $P(\X | I)$ is well defined. We may then define the updated joint distribution
    \begin{equation*}
        P_{\text{new}}(\X) = P(Y | \theta)P(\theta | I),
    \end{equation*}
    and the updated marginal (predictive) distribution for $Y$ becomes
    \begin{equation*}
        P_{\text{new}}(Y) = \int P(Y | \theta) dP(\theta | I).
    \end{equation*}
\end{ex}
\medskip
Note that the distribution $P_{\text{new}}(\X)$ in \exref{ex:X_theta_2} is obtained by only updating the belief with respect to epistemic uncertainty, 
and that 
\begin{equation*}
    P_{\text{new}}(\X) \neq P(\X | I) = P(Y|I, \theta)P(\theta|I).
\end{equation*}
For instance, if $I$ corresponds to an observation of $Y$, e.g. $I = \{Y = y \}$, then $P(Y|I) = \delta(y)$, the Dirac delta at $y$, 
whereas $P(\theta | I)$ is the updated distribution for $\theta$ having observed one realization of $Y$.
In the following, we will refer to the kind of Bayesian updating in \exref{ex:X_theta_2} as \emph{epistemic updating}.

This epistemic updating of the model considered in \exref{ex:X_theta} and \exref{ex:X_theta_2} should be fairly intuitive, if 
\begin{enumerate}
    \item All epistemic uncertainty is represented by a single parameter $\theta$, and
    \item{$\theta$ is a familiar object like a number or a vector in $\mathbb{R}^n$.}
\end{enumerate}

But what can we say in a more general setting? It is common that epistemic uncertainty comes from lack of knowledge related to \emph{functions}. This is the case with probabilistic emulators and surrogate models. The input to these functions may contain epistemic and/or aleatory uncertainty as well. 
Can we talk about isolating and modifying the epistemic uncertainty in such a model, without making reference to the specific details of how the model has been created?
In the following we will show that with the measure-theoretic framework, we can still make use of a simple formulation like the one in \exref{ex:X_theta_2}.

\subsection{The probability space}
\label{sec:probspace}
Let $\X$ be a random variable containing both aleatory and epistemic uncertainty. 
In order to describe how $\X$ can be treated like in \exref{ex:X_theta} and \exref{ex:X_theta_2}, but for the general setting, 
we will first recall some of the basic definitions from measure theory and measure-theoretic probability. 

To say that $\X$ is a \emph{random variable}, means that $\X$ is defined on some \emph{measurable space} $(\Omega, \mathcal{F})$. 
Here, $\Omega$ is a set, and if $\X$ takes values in $\mathbb{R}^n$ (or some other measurable space), 
then $\X$ is a so-called \emph{measurable function}, $\X(\omega) : \Omega \rightarrow \mathbb{R}^n$ (to be defined precisely later). 
Any randomness or uncertainty about $\X$ is just a consequence of uncertainty regarding $\omega \in \Omega$. 
As an example, $\X$ could relate to a some 1-year extreme value, whose uncertainty comes from day to day fluctuations, or some 
fundamental stochastic phenomenon represented by $\omega \in \Omega$. Examples of natural sources of uncertainty are weather or human actions in large scale. Therefore, whether modeling weather, option prices, structural safety at sea or traffic networks, stochastic models should be used.

The probability of the event $\{ \X \in E \}$, for some subset $E \subset \mathbb{R}^n$, is really the probability of $\{ \omega \in \X^{-1}(E) \}$. 
Technically, we need to ensure that $\{ \omega \in \X^{-1}(E) \}$ is something that we can compute the probability of, 
and for this we need $\mathcal{F}$. $\mathcal{F}$ is a collection of subsets of $\Omega$, and represents \emph{all possible events} (in the "$\Omega$-world").
When $\mathcal{F}$ is a $\sigma$\emph{-algebra}\footnote{
This means that
1) $\Omega \in \mathcal{F}$, 2) if $S \in \mathcal{F}$ then also the complement $\Omega \setminus S \in \mathcal{F}$, 
and 3) if $S_1, S_2, \cdots $ is a countable set of events then also the union $S_1 \cup S_2 \cup \dots$ is in $\mathcal{F}$. 
Note that if these properties hold, many other types of events (e.g. countable intersections) will have to be included as a consequence.
}
the pair $(\Omega, \mathcal{F})$ becomes a \emph{measurable space}.

So, when we define $\X$ as a \emph{random variable} taking values in $\mathbb{R}^n$, this means that there exists some \emph{measurable space} $(\Omega, \mathcal{F})$, such that any event $\{ \X \in E \}$ in the "$\mathbb{R}^n$-world" (which has its own $\sigma$-algebra) has 
a corresponding event $\{ \omega \in \X^{-1}(E) \} \in \mathcal{F}$ in the "$\Omega$-world".
It also means that we can define a probability measure on $(\Omega, \mathcal{F})$ that gives us the probability of each event, 
but before we introduce any specific probability measure, $\X$ will just be a \emph{measurable function}\footnote{
    By definition, given two measure spaces $(\Omega, \mathcal{F})$ and $(\mathbb{X}, \mathcal{X})$, the function $\X : \Omega \rightarrow \mathbb{X}$ 
    is measurable if and only if $\X^{-1}(A) \in \mathcal{F} \ \forall A \in \mathcal{X}$.
}.
\begin{itemize}
    \item[-] We start with assuming that there exists some \emph{measurable space} $(\Omega, \mathcal{F})$ where 
    $\X$ is a \emph{measurable function}.
\end{itemize}
The natural way to make $\X$ into a random variable is then to introduce some probability measure\footnote{
    A function $P : \mathcal{F} \rightarrow [0, 1]$ such that 1) $P(\Omega) = 1$ and 2) 
    $P(\cup E_i) = \sum P(E_i)$ for any countable collection of pairwise disjoint events $E_i$.}
$P$ on $\mathcal{F}$, giving us the \emph{probability space} $(\Omega, \mathcal{F}, P)$.
\begin{itemize}
    \item[-] Given a \emph{probability measure} $P$ on $(\Omega, \mathcal{F})$ we obtain the \emph{probability space} $(\Omega, \mathcal{F}, P)$ on which $\X$ is defined
    as a \emph{random variable}. 
\end{itemize}
We have considered here, for familiarity, that $\X$ takes values in $\mathbb{R}^n$. 
When no measure and $\sigma$-algebra is stated explicitly, one can assume that $\mathbb{R}^n$ is endowed with the Lebesgue measure 
(which underlies the standard notion of length, area and volume etc.) and the Borel $\sigma$-algebra (the smallest $\sigma$-algebra containing all open sets).
Generally, $\X$ can take values in any measurable space. For example, 
$\X$ can map from $\Omega$ to a space of functions. This is important in the study of stochastic processes. 

\subsection{The epistemic sub $\sigma$-algebra $\mathcal{E}$}
In the probability space $(\Omega, \mathcal{F}, P)$, recall that the $\sigma$-algebra $\mathcal{F}$ contains all possible events. 
For any random variable $\X$ defined on $(\Omega, \mathcal{F}, P)$, the knowledge that some event has occurred provides information 
about $\X$. This information may relate to $\X$ in a way that it only affects epistemic uncertainty, only aleatory uncertainty, or both. 
We are interested in specifying the events $e \in \mathcal{F}$ that are associated with epistemic information alone. 
It is the probability of these events we want to update as new information is obtained. 
The collection $\mathcal{E}$ of such sets is itself a $\sigma$-algebra, and we say that 
\begin{equation}
    \label{eq:epistemic_sigma_alg}
    \mathcal{E} \subseteq \mathcal{F}
\end{equation}
is the sub $\sigma$-algebra of $\mathcal{F}$ representing epistemic information. 

We illustrate this in the following examples. In Example \ref{ex:coin_1}, we consider the simplest possible scenario represented by the flip of a biased coin, 
and in Example \ref{ex:uq_1} a  
familiar scenario from uncertainty quantification involving uncertainty with respect to functions. 
\medskip
\begin{ex}(Coin flip)
    \label{ex:coin_1}\\
    \noindent Define $\X = (Y, \theta)$ as in \exref{ex:X_theta}, and let $Y \in \{ 0, 1 \}$ denote the outcome of a coin flip where "heads" is represented by $Y = 0$ and "tails" by $Y = 1$.
    Assume that $P(Y = 0) = \theta$ for some fixed but unknown $\theta \in [0, 1]$. For simplicity we assume that 
    $\theta$ can only take two values, $\theta \in \{ \theta_1, \theta_2 \}$ (e.g. there are two coins but we do not know which one is being used).
    
    Then $\Omega = \{ 0, 1 \} \times \{ \theta_1, \theta_2 \}$, $\mathcal{F} = 2^\Omega$ and 
    $\mathcal{E} = \{ \emptyset, \Omega, \{ (0, \theta_1), (1, \theta_1) \}, \{ (0, \theta_2), (1, \theta_2) \} \}$.
\end{ex}

\begin{ex}(UQ)
\label{ex:uq_1}\\
    \noindent Let $\X = (\x, \y)$ where $\x$ is an aleatory random variable, and $\y$ is the result of a fixed but unknown function applied to $\x$. 
    We let $\y = \hat{f}(\x)$ where $\hat{f}$ is a function-valued epistemic random variable. 
    
    If $\x$ is defined on a probability space $(\Omega_\x, \mathcal{F}_\x, P_\x)$ and
    $\hat{f}$ is a stochastic process defined on $(\Omega_f, \mathcal{F}_f, P_f)$, then 
    $(\Omega, \mathcal{F}, P)$ can be defined as the product of the two spaces 
    and $\mathcal{E}$ as the projection $\mathcal{E} = \{ \Omega_\x \times A \mid A \in \mathcal{F}_f \}$.
\end{ex}

\medskip

In the following, we assume that the epistemic sub $\sigma$-algebra $\mathcal{E}$ has been identified. 

Given a random variable $\X$, we say that $\X$ is $\mathcal{E}$-measurable if $\X$ is measurable as a function defined on $(\Omega, \mathcal{E})$. 
We say that $\X$ is independent of $\mathcal{E}$, if the conditional probability $P(\X | e)$ is equal to $P(\X)$ for any event $e \in \mathcal{E}$.
With our definition of $\mathcal{E}$, we then have for any random variable $\X$ on $(\Omega, \mathcal{F}, P)$ that
\begin{itemize}
    \item[-] $\X$ is purely epistemic if and only if $\X$ is $\mathcal{E}$-measurable,
    \item[-] $\X$ is purely aleatory if and only if $\X$ is independent of $\mathcal{E}$.
\end{itemize}

\subsection{Epistemic conditioning}
\label{sec:epistemic_conditioning}
Let $\X$ be a random variable on $(\Omega, \mathcal{F}, P)$ that may contain both epistemic and aleatory uncertainty, 
and assume that the epistemic sub $\sigma$-algebra $\mathcal{E}$ is given. 
By epistemic conditioning, we want to update the epistemic part of the uncertainty in $\X$ using some set of information $I$. 
In \exref{ex:coin_1} this means updating the probabilities $P(\theta = \theta_1)$ and $P(\theta = \theta_2)$, 
and in \exref{ex:uq_1} this means updating $P_f$. 
In order to achieve this in the general setting, we first need a way to decouple epistemic and aleatory uncertainty. 
This can actually be made fairly intuitive, if we rely on the following assumption:
\begin{assumption}
\label{assumption}
    There exists a random variable $\theta : \Omega \rightarrow \Theta$ that generates\footnote{
    There exists some measurable space $(\Theta, \mathcal{T})$ and a $\mathcal{F}$-measurable function $\theta :  \Omega \rightarrow \Theta$ such 
    that $\mathcal{E} = \sigma(\theta)$, the smallest $\sigma$-algebra containing all of the sets $\theta^{-1}(T)$ for $T \in \mathcal{T}$.
    } 
    $\mathcal{E}$.
\end{assumption}
If this generator $\theta$ exists, then for any fixed value $\theta \in \Theta$, we have that $\X | \theta$ is independent of $\mathcal{E}$. 
Hence $\X | \theta$ is purely aleatory and $\theta$ is purely epistemic. 

We will call $\theta$ the \emph{epistemic generator}, and we can interpret $\theta$ as a signal that reveals all epistemic information when known. 
That is, if $\theta$ could be observed, then knowing the value of $\theta$ would remove all epistemic uncertainty from our model. 
As it turns out, under fairly mild conditions one can always assume existence of this generator. 
One sufficient condition is that $(\Omega, \mathcal{F}, P)$ is a \emph{standard} probability space, and then the statement holds up to sets of measure zero. 
This is a technical requirement to avoid pathological cases, and does not provide any new intuition that we see immediately useful, so we postpone further explanation 
to \appref{app:epiestemic_gen_proof}. 
\medskip
\begin{ex}(Coin flip -- epistemic generator)
    \label{ex:coin_2}\\
    \noindent In the coin flip example, the variable $\theta \in \{ \theta_1, \theta_2 \}$ which generates $\mathcal{E}$ is already specified. 
\end{ex}
    
\begin{ex}(UQ -- epistemic generator)
    \label{ex:uq_2}\\
    \noindent In this example, when $(\Omega, \mathcal{F}, P)$ is the product of an aleatory space $(\Omega_\x, \mathcal{F}_\x, P_\x)$ and 
    an epistemic space $(\Omega_f, \mathcal{F}_f, P_f)$, we could let $\theta : \Omega = \Omega_\x \times \Omega_f \rightarrow \Omega_f$ be the 
    projection $\theta(\omega_\x, \omega_f) = \omega_f$. 
    
    Alternatively, given only the space $(\Omega, \mathcal{F}, P)$ where both $\x$ and $\hat{f}$ are defined, 
    assume that $\hat{f}$ is a Gaussian process (or some other stochastic process for which the Karhunen–Lo\'eve theorem holds). 
    Then there exists a sequence of deterministic functions $\phi_1, \phi_2, \dots$ and
    an infinite-dimensional variable $\theta = (\theta_1, \theta_2, \dots)$ such that 
    $\hat{f}(\x) = \sum_{i=1}^{\infty} \theta_i \phi_i(\x)$, and we can let $\mathcal{E}$ be generated by $\theta$.
\end{ex}
\medskip
The decoupling of epistemic and aleatory uncertainty is then obtained by considering the joint variable $(\X, \theta)$ instead of $\X$ alone, because
\begin{equation}
    \label{eq:joint_prob}
    P(\X, \theta) = P(\X \mid \theta)P(\theta).
\end{equation}
From \eqnref{eq:joint_prob} we see how the probability measure $P$ becomes the product of the epistemic probability $P(\theta)$ and 
the aleatory probability $P(\X | \theta)$ when applied to $(\X, \theta)$.

Given new information, $I$, we will update our beliefs about $\theta$, $P(\theta) \rightarrow P(\theta | I)$, and we define the epistemic conditioning as follows:
\begin{equation}
    P_{\text{new}}(\X, \theta) = P(\X \mid \theta)P(\theta \mid I).
\end{equation}

\subsection{Two types of assumptions}
Consider the probability space $(\Omega, \mathcal{F}, P)$, with epistemic sub $\sigma$-algebra $\mathcal{E}$. 
Here $\mathcal{E}$ represents \emph{epistemic information}, which is the information associated with \emph{assumptions}.
In other words, an epistemic event $e \in \mathcal{E}$ represents an assumption. 
In fact, given a class of assumptions, the following Remark \ref{remark: sigma-alg}, shows why $\sigma$-algebras are appropriate structures. 
\begin{remark}
\label{remark: sigma-alg}
    Let $\mathcal{E}$ be a collection of assumptions. If $e \in \mathcal{E}$, this means that it is possible to assume that $e$ is true. 
    If it is also possible to assume that that $e$ is false, then. $\Bar{e} \in \mathcal{E}$ as well. It may then also be natural to require 
    that $e_1, e_2 \in \mathcal{E} \Rightarrow e_1 \cap e_2 \in \mathcal{E}$, and so on. These are the defining properties of a $\sigma$-algebra.
\end{remark}

For any random variable $\X$ defined on $(\Omega, \mathcal{F}, P)$, when $\mathcal{E}$ is a sub $\sigma$-algebra of $\mathcal{F}$, $\X | e$ for $e \in \mathcal{E}$ is well defined, and represents the random variable under the assumption $e$.
In particular, given any fixed \emph{epistemic event} $e \in \mathcal{E}$ we have a corresponding \emph{aleatory distribution} $P(\X | e)$ over $\X$,  
and the conditional $P(\X | \mathcal{E})$ is the random measure corresponding to $P(\X | e)$ when $e$ is a random epistemic event in $\mathcal{E}$.
Here, the global probability measure $P$ when applied to $e$, $P(e)$, is the belief that $e$ is true. In \secref{sec:epistemic_conditioning} 
we discussed updating the part of $P$ associated with epistemic uncertainty. We also introduced the epistemic generator $\theta$ in order to associate the event $e$ with an outcome $\theta(e)$, and make use of $P(\X | \theta)$ in place of $P(\X | \mathcal{E})$. 
This provides a more intuitive interpretation of the assumptions that are \emph{measurable}, i.e. those whose belief we may specify through $P$.

Of course, the measure $P$ is also based on assumptions.  
For instance, if we in \exref{ex:X_theta} assume that $Y$ follows a \emph{normal} distribution. 
One could in principle specify a (measurable) space of probability distributions, from which the normal distribution is one example.
Otherwise, we view the normality assumption as a \emph{structural} assumption related to the probabilistic model for $\X$, i.e. the measure $P$. These kinds of assumptions cannot be treated the same way as assumptions related to measurable events. 
For instance, the consequence of the complement assumption \textit{"$Y$ does not follow a normal distribution"} is not well defined. 

In order to avoid any confusion, we split the assumptions into two types:

\begin{enumerate}
    \item The measurable assumptions represented by the $\sigma$-algebra $\mathcal{E}$, and
    \item the set $M$ of structural assumptions underlying the probability measure $P$. 
\end{enumerate} 

This motivates the following definition.
\begin{defn}[Structural assumptions]
    We let $M$ denote the set of structural assumptions that defines a probability measure on $(\Omega, \mathcal{F})$,
    which we may write $P_M(\cdot)$ or $P(\cdot \mid M)$. 
\end{defn}
\medskip
We may also refer to $M$ as the \emph{non-measurable} assumptions, to emphasize that $M$ contains all the assumptions not 
covered by $\mathcal{E}$.
When there is no risk of confusion we will also suppress the dependency on $M$ and just write $P(\cdot)$.
Stating the set $M$ explicitly is typically only relevant for scenarios where we consider changes being made to the actual system that is being modelled, 
or for evaluating different candidate models, e.g. through the marginal likelihood $P(I | M)$.
In practice one would also state $M$ so that decision makers can 
determine their level of trust in the probabilistic model, and the appropriate level of caution when applying the model.

As we will see in the upcoming section, making changes to $M$ and making changes to how $P_M$ acts on events in $\mathcal{E}$ are 
the two main ways in which we update a \emph{probabilistic digital twin}.

\section{The Probabilistic Digital Twin}
\label{sec:PDT}
The object that we will call \emph{probabilistic digital twin}, PDT for short, is a probabilistic model of a physical system.
It is essentially a (possibly degenerate) probability distribution of a vector $\X$, representing the relevant attributes of the system, 
but where we in addition require the specification of epistemic uncertainty (assumptions) and how this uncertainty may be updated given new information. 

Before presenting the formal definition of a probabilistic digital twin, we start with an example showing why the identification of epistemic uncertainty is important.

\subsection{Why distinguish between aleatory and epistemic uncertainty?}
The decoupling of epistemic and aleatory uncertainty (as described in \secref{sec:epistemic_conditioning}) is central in the PDT framework. There are two good reasons for doing this:

\begin{enumerate}
    \item We want to make decisions that are \emph{robust with respect to epistemic uncertainty}. 
    \item We want to study the effect of gathering information. 
\end{enumerate}

Item 1. relates to the observation that decision theoretic approaches based on expectation may not be robust. That is, 
if we marginalize out the epistemic uncertainty (and considering only $E_{\theta}[P(\X | \theta)] = \int P(\X | \theta) dP_\theta$). 
We give two examples of this below, see Example \ref{ex: coin_robust} and Example \ref{ex: UQ_robust}. 

Item 2. means that by considering the effect of information on epistemic uncertainty, we can evaluate the 
value of gathering information. This is discussed in further detail in \secref{sec:optimal_info_gathering}.
\medskip
\begin{ex}(Coin flip -- robust decisions)\\
    Continuing from the coin flip example (see Example \ref{ex:coin_1}), we let $\theta_1 = 0.5, \theta_2 = 0.99$. Assume that you are given the option to guess the outcome of $\X$.
    If you guess correct, you collect a reward of $R = 10^6 \ \$$, otherwise you have to pay $L = 10^6 \ \$$. 
    A priori your belief about the bias of the coin is that $P(\theta = 0.5) = P(\theta = 0.99) = 0.5$. 
    If you consider betting on $\X = 0$, then the expected return, obtained by marginalizing over $\theta$, becomes
    $P(\theta = 0.5)(0.5R - 0.5L) + P(\theta = 0.99)(0.99 R - 0.01L) = 490.000 \$$. 
    
    This is a scenario where decisions supported by taking the expectation with respect to epistemic uncertainty is not robust, 
    as we believe that $\theta = 0.5$ and $\theta = 0.99$ are equally likely, and if $\theta = 0.5$ we will lose $10^6 \ \$$
    $50 \%$ of the time by betting on $\X = 0$.
    \label{ex: coin_robust}
\end{ex}

\begin{ex}(UQ -- robust decisions)\\
    This example is a continuation of Example \ref{ex:uq_1} and Example \ref{ex:uq_2}.

    In structural reliability analysis, we are dealing with an unknown function $g$ with the property that 
    the event $\{ y = g(\x) < 0 \}$ corresponds to failure. When $g$ is represented by a random function $\hat{g}$ with epistemic uncertainty, 
    the failure probability is also uncertain. Or in other words, if $\hat{g}$ is epistemic then $\hat{g}$ is a function 
    of the generator $\theta$. Hence, the failure probability is a function of $\theta$. We want to make use of a conservative estimate of
    the failure probability, i.e., use a conservative value of $\theta$. $P(\theta)$ tells us how conservative a given value of $\theta$ is. 
    \label{ex: UQ_robust}
\end{ex}

\subsection{The attributes $\X$}
To define a PDT, we start by considering a vector $\X$ consisting of the \emph{attributes} of some system. 
This means that $\X$ is a representation of the physical object or asset that we are interested in. In general, $\X$ describes the physical \emph{system}.
In addition, $\X$ must contain attributes related to any type of information that we want to make use of. 
For instance, if the information consists of observations, the relevant observable quantities, as well as 
attributes related to measurement errors or noise, may be included in $\X$. 
In general, we will think of a \emph{model} of a system as a set of assumptions that describes 
how the components of $\X$ are related or behave. 
The canonical example here is where some physical quantity is inferred from observations including errors and noise, 
in which case a model of the physical quantity (physical system) is connected with a model of the data generating process (observational system).  
We are interested in modelling dependencies with associated uncertainty related to the components of $\X$, and treat $\X$ as a random variable.

The attributes $\X$ characterise the state of the system and the processes that the PDT represents. $\X$ may for instance include:
\begin{itemize}
    \item System parameters \textit{representing quantities that have a fixed, but possibly uncertain, value. For instance, these parameters may be related to the system configuration.}
    \item System variables \textit{that may vary in time, and which value may be uncertain.} 
    \item System events \textit{i.e., the occurrence of defined state transitions.}
\end{itemize}
In risk analysis, one is often concerned with risk measures given as quantified properties of $\X$, usually in terms of expectations. 
For instance, if $\X$ contains some extreme value (e.g. the 100-year wave) or some specified event of failure (using a binary variable), 
the expectations of these may be compared against risk acceptance criteria to determine compliance.

\subsection{The PDT definition}
Based on the concepts introduced so far, we define the PDT as follows:

\begin{defn}[Probabilistic Digital Twin]
\label{def: PDT}
    A Probabilistic Digital Twin (PDT) is a triplet
    $$(\X, A, I)$$ 
    where $\X$ is a vector of attributes of a system, 
    $A$ contains the \underline{assumptions} needed to specify a probabilistic model, and $I$ contains \underline{information} regarding actions and 
    observations:\\

    $A = ((\Omega, \mathcal{F}), \mathcal{E}, M)$, where $(\Omega, \mathcal{F})$ is a measure space where $\X$ is measurable, 
    and $\mathcal{E}$ is the sub $\sigma$-algebra representing epistemic information.
    $M$ contains the structural assumptions that defines a probability measure $P_M$ on $(\Omega, \mathcal{F})$. \\ 
    
    $I$ is a set consisting of events of the form $(d, o)$, 
    where $d$ encodes a description of the conditions under which the observation $o$ was made, 
    and where the likelihood $P(o | \X, d)$ is well defined. 
    For brevity, we will write this likelihood as $P(I | \X)$ when $I$ contains multiple events of this sort. 
\end{defn}
When $M$ is understood, and there is no risk on confusion, we will drop stating the dependency on $M$ explicitly and just refer to the probability space $(\Omega, \mathcal{F}, P)$.

It is important to note that consistency between $I$ and $P(\X)$ is required. That is, when using the probabilistic model for $\X$, it should be possible 
to simulate the type of observations given by $I$. In this case the likelihood $P(I | \X)$ is well defined, and the epistemic updating of $\X$ can 
be obtained from Bayes' theorem.

Finally, we note that with this setup the information $I$ may contain observations made under different conditions than what is currently 
specified through $M$. The information $I$ is generally defined as a set of events, given as 
pairs $(d, o)$, where the $d$ encodes the relevant action leading to observing $o$, as well as a description of the conditions under which $o$ was observed. 
Here $d$ may relate to modifications of the structural assumptions $M$, for instance if the the causal relationships that describes the model of $\X$ under observation of $o$ is not the same as what is currently represented by $M$. This is the scenario when we perform controlled experiments. 
Alternatively, $(d, o)$ may represent a passive observation, e.g. $d = $ \textit{"measurement taken from sensor 1 at time 01:02:03"}, $o = 1.7$ mm. 
We illustrate this in the following example.

\begin{ex}(Parametric regression)
\label{ex:param_reg}\\
    \noindent Let $(x_1, x_2)$ denote two physical quantities where $x_2$ depends on $x_1$, and let $(y, \varepsilon)$ represent an observable quantity where 
    $y$ corresponds to observing $x_2$ together with additive noise $\varepsilon$. Set $\X = (x_1, x_2, y, \varepsilon)$.
    
    \begin{figure}[h]
    \centering
        \begin{tikzpicture}
            \tikzstyle{roundnode} = [circle, minimum size=0.8cm, text centered, draw=black]
            \tikzstyle{arrow} = [thick,->,>=stealth]

            \node (x1) [roundnode] {$x_1$};
            \node (x2) [roundnode, right of=x1, xshift=1cm] {$x_2$};
            \node (y) [roundnode, right of=x2, xshift=1cm] {$y$};
            \node (eps) [roundnode, right of=y, xshift=1cm] {$\varepsilon$};

            \draw [arrow] (x1) -- (x2);
            \draw [arrow] (x2) -- (y);
            \draw [arrow] (eps) -- (y);

        \end{tikzpicture}
    \caption{A standard regression model as a PDT.}
    \label{fig:ex_regression_graph}
    \end{figure}
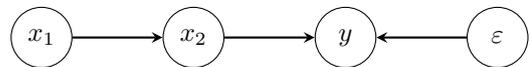
    We define a model $M$ corresponding to $x_1 \sim p_{x_1}(x_1 | \theta_1)$, $x_2 = f(x_1, \theta_2)$, $y = x_2 + \varepsilon$ and $\varepsilon \sim p_{\varepsilon}$, 
    where $p_{x_1}$ is a probability density depending on the parameter $\theta_1$ and $f(\cdot, \theta_2)$ is a deterministic function depending on the parameter $\theta_2$.
    
    $\theta_1$ and $\theta_2$ are epistemic parameters for which we define a joint density $p_{\theta}$.
    
    Assume that $I = \{ (d^{(1)}, o^{(1)}), \dots, (d^{(n)}, o^{(n)}) \}$ is a set of controlled experiments, where 
    $d^{(i)} = (\text{set } x_1 = x_{1}^{(i)})$ and $o^{(i)}$ is a corresponding observation of $y|(x_1 = x_{1}^{(i)}, \varepsilon = \varepsilon^{(i)})$ for a selected set 
    of inputs $x_{1}^{(i)}, \dots, x_{1}^{(n)}$ and unknown i.i.d. $\varepsilon^{(i)} \sim p_\varepsilon$. In this scenario, regression is performed by updating the distribution $p_\theta$ 
    to agree with the observations: 
    \begin{equation}
        \label{eq:parem_reg}
        \begin{split}
            &p_{\theta}(\theta | I) = p_{\theta}(\theta_1 | \theta_2) p_{\theta}(\theta_2 | I) \\
            &= \frac{1}{Z} \displaystyle \left(  \prod_{i=1}^{n} p_{\varepsilon}\left( o^{(i)} - f(x_{1}^{(i)}, \theta_2) \right)  \right) p_{\theta}(\theta),    
        \end{split}
    \end{equation}
    where $Z$ is a constant ensuring that the updated density integrates to one.
    
    If instead $I$ corresponds to direct observations, $d^{(i)} = (\text{observe } y^{(i)})$, $o^{(i)} = y^{(i)}$, then $p_{\theta}(\theta | I)$ corresponds to 
    using $x_1$ instead of $x_{1}^{(i)}$ and multiplying with $p_{x_1}(x_1|\theta_1)$ in \eqnref{eq:parem_reg}.
\end{ex}

Note that the scenario with controlled experiments in \exref{ex:param_reg} corresponds to a different model than the one in \figref{fig:ex_regression_graph}.
This is a familiar scenario in the study of causal inference, where actively setting the value of $x_1$ is the do-operator (see Pearl \cite{Pearl:1995:causal_diagrams}) 
which breaks link between $x_1$ and $x_2$.

\subsection{Corroded pipeline example}
\label{sec:corr_pipeline}
To give a concrete example of a system where the PDT framework is relevant, we consider the following model from Agrell and Dahl \cite{Agrell:2020:DOE_SRA}.
This is based on a probabilistic structural reliability model which is recommended for engineering assessment of offshore pipelines with corrosion (DNV GL RP-F101 \citep{RP-F101}).
It is a model of a physical failure mechanism called pipeline burst, which may occur
when the pipeline's ability to withstand the high internal pressure has been reduced as a consequence of corrosion.
We will describe just a general overview of this model, and refer to \citep[][Example 4]{Agrell:2020:DOE_SRA} for specific details regarding probability distributions etc. 
Later, in \secref{sec:corroded_pipeline_RL}, we will revisit this example and make use of reinforcement learning to search for an optimal way of updating the PDT. 

\begin{figure}[h]
    \centering
        \begin{tikzpicture}[scale=0.9, every node/.style={scale=0.9}]
            \tikzstyle{roundnode} = [circle, minimum size=0.9cm, text centered, draw=black]
            \tikzstyle{roundnode_fill} = [circle, minimum size=0.9cm, text centered, draw=black, fill = black!15]
            \tikzstyle{arrow} = [thick,->,>=stealth]

            \node (D) [roundnode] {$D$};
            \node (t) [roundnode, right of=D, xshift=0.5cm] {$t$};
            \node (UT) [roundnode, right of=t, xshift=0.5cm] {$s$};
            \node (dt) [roundnode_fill, right of=UT, xshift=0.75cm] {$d$};
            \node (l) [roundnode, right of=dt, xshift=0.5cm] {$l$};
            
            \node (mod_sig) [roundnode, below of=D, yshift=-1.0cm, xshift=-0.0cm] {$\sigma_{\text{m}}$};
            \node (mod_mu) [roundnode_fill, left of=mod_sig, xshift=-0.5cm] {$\mu_{\text{m}}$};
            \node (Xm) [roundnode, below of=mod_sig, yshift=-0.5cm] {$X_{\text{m}}$};
            
            \node (Pcap_FE) [roundnode_fill, below of=UT, yshift=-1.0cm] {$p_{\text{FE}}$};
            \node (Pcap) [roundnode, below of=Pcap_FE, yshift=-0.5cm] {$p_{c}$};
            \node (pd) [roundnode, right of=Pcap, xshift=1.5cm] {$p_{d}$};
            \node (g) [roundnode, below of=Pcap, yshift=-0.5cm, xshift = 1.5cm] {$g$};
            
            \draw [arrow] (D) -- (Pcap_FE);
            \draw [arrow] (t) -- (Pcap_FE);
            \draw [arrow] (UT) -- (Pcap_FE);
            \draw [arrow] (dt) -- (Pcap_FE);
            \draw [arrow] (l) -- (Pcap_FE);
            \draw [arrow] (mod_sig) -- (Xm);
            \draw [arrow] (mod_mu) -- (Xm);
            \draw [arrow] (Xm) -- (Pcap);
            \draw [arrow] (Pcap_FE) -- (Pcap);
            \draw [arrow] (Pcap) -- (g);
            \draw [arrow] (pd) -- (g);
            
            \node [above of = t, yshift = -0.25cm,] {\footnotesize{\textbf{Pipeline}}};
            \node [above of = dt, yshift = -0.25cm, xshift=0.8cm] {\footnotesize{\textbf{Defect}}};
            \node [above of = mod_mu, yshift = -0.25cm, xshift=0.8cm] {\footnotesize{\textbf{Model discrepancy}}};

            \node [above of = pd, yshift = -0.25cm, xshift = 0.0cm] {\footnotesize{\textbf{Load}}};
            \node [above of = Pcap, yshift = -0.25cm, xshift = -0.8cm, text width=1cm, align=center, rotate=90] {\footnotesize{\textbf{Capacity}}};

        \end{tikzpicture}
    \caption{Graphical representation of the corroded pipeline structural reliability model. 
    The shaded nodes $d$, $p_{\text{FE}}$ and $\mu_{\text{m}}$ have associated epistemic uncertainty.}
    \label{fig:corrpipe_problem}
\end{figure}
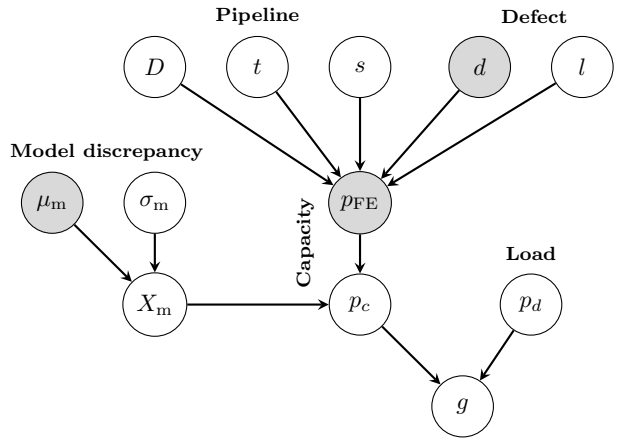

\figref{fig:corrpipe_problem} shows a graphical representation of the structural reliability model.
Here, a steel \textbf{pipeline} is characterised by the outer diameter $D$, the wall thickness $t$
and the ultimate tensile strength $s$. 
The pipeline contains a rectangular shaped \textbf{defect} with a given depth $d$ and length $l$.
Given a pipeline $(D, t, s)$ with a defect $(d, l)$, we can determine the pipeline's pressure 
resistance \textbf{capacity} (the maximum differential pressure the pipeline can withstand before bursting). 
We let $p_{\text{FE}}$ denote the capacity coming from a Finite Element simulation of the physical phenomenon. 
From the theoretical capacity $p_{\text{FE}}$, we model the true pipeline capacity as a function of $p_{\text{FE}}$ and $X_{\text{m}}$,
where $X_{\text{m}}$ is the \textbf{model discrepancy}.
For simplicity we have assumed that $X_{\text{m}}$ does not depend on the type of pipeline and defect, 
and we will also assume that $\sigma_{\text{m}}$ is fixed, and only the mean $\mu_{\text{m}}$ can be inferred from observations. 
Finally, given the pressure \textbf{load} $p_d$, the limit state representing the transition to failure is then given as $g = p_c - p_d$, 
and the probability of failure is defined as $P(g \leq 0)$.

If we let $\X$ be the random vector containing all of the nodes in \figref{fig:corrpipe_problem}, then 
$\X$ represents a probabilistic model of the physical system. 
In this example, we want to model some of the uncertainty related to the defect size, the model uncertainty, and the capacity as epistemic. 
We assume that the defect depth $d$ has a fixed but unknown value, that can be inferred through observations that include noise.
Similarly, the model uncertainty $X_{\text{m}}$ can be determined from experiments. 
Uncertainty with respect to $p_{\text{FE}}$ comes from the fact that evaluating the true value of $p_{\text{FE}} | (D, t, s, d, l)$ involves 
a time-consuming numerical computation. Hence, $p_{\text{FE}}$ can only be known for a finite, and relatively small set of input combinations. 
We can let $\hat{p}_{\text{FE}}$ denote a stochastic process that models our uncertainty about $p_{\text{FE}}$. 
To construct a PDT from $\X$ we will let $\hat{p}_{\text{FE}}$ take the place of $p_{\text{FE}}$, and specify that 
$d, \mu_{\text{m}}$ and $\hat{p}_{\text{FE}}$ are epistemic, i.e. $\mathcal{E} = \sigma(d, \mu_{\text{m}}, \hat{p}_{\text{FE}})$.

If we want a way to update the epistemic uncertainty based on observations, we also need to specify the relevant data generating process. 
In this example, we assume that there are three different ways of collecting data:
\begin{enumerate}
    \item \textbf{Defect measurement:} We assume that noise perturbed observations of the relative depth, $d/t + \varepsilon$, can be made.
    \item \textbf{Computer experiment:} Evaluate $p_{\text{FE}}$ at some selected input $(D, t, s, d, l)$.
    \item \textbf{Lab experiment:} Obtain one observation of $X_{\text{m}}$.
\end{enumerate}

As the defect measurements requires specification of an additional random variable, we have to include $\varepsilon$ or 
$(d/t)_{\text{obs}} = d/t + \varepsilon$ in $\X$ as part of the complete probabilistic model. 
This would then define a PDT where epistemic updating is possible. 

The physical system that the PDT represents in this example is rarely viewed in isolation. 
For instance, the random variables representing the pipeline geometry and material are the result of uncertainty or variations 
in how the pipeline has been manufactured, installed and operated. And the size of the defect is the result of a chemical process, 
where scientific models are available. It could therefore be natural to view the PDT from this example as a component of a bigger PDT, where 
probabilistic models of the manufacturing, operating conditions and corrosion process etc. are connected. 
This form of modularity is often emphasized in the discussion of digital twins, 
and likewise for the kind of Bayesian network type of models as considered in this example.

\section{Sequential decision making}
\label{sec:seq_dec_making}
We now consider how the PDT framework may be adopted in real-world applications. 
As with any statistical model of this form, the relevant type of applications are related to prediction and inference. 
Since the PDT is supposed to provide a one-to-one correspondence (including uncertainty) with a real physical system, 
we are interested in using the PDT to understand the consequences of actions that we have the option to make. 
In particular, we will consider the discrete sequential decision making scenario, where get the opportunity to make a decision, 
receive information related to the consequences of this decision, and use this to inform the next decision, and so on.

\begin{figure}[h]
\centering
    \begin{tikzpicture}
    
        \tikzstyle{roundnode} = [circle, minimum size=1.4cm, text centered, draw=black]
    
        \def \radius {3cm}
        \def \margin {15} 
        
        \node (1) [] at ({360/3 * (1 - 1)}:\radius)  {\HUGE \faGlobe};
        \node (2) [roundnode] at ({360/3 * (2 - 1)}:\radius) {PDT};
        \node (3) [roundnode] at ({360/3 * (3 - 1)}:\radius) {\parbox{0.8cm}{ \scriptsize \centering
                $\Delta M$ \\
                $\Delta I$ }};
    
        \draw[<-, >=latex, postaction={decorate,decoration={text effects along path,
        text={\ Apply policy\ }, text align=center, reverse path,
        text effects/.cd, 
          text along path, 
          every character/.style={fill=white, yshift=-0.5ex}}}] ({360/3 * (1 - 1)+\margin}:\radius) arc ({360/3 * (1 - 1)+\margin}:{360/3 * (1)-\margin}:\radius);
    
        \draw[<-, >=latex, postaction={decorate,decoration={text effects along path,
        text={\ Update model\ }, text align=center, reverse path,
        text effects/.cd, 
          text along path, 
          every character/.style={fill=white, yshift=-0.5ex}}}] ({360/3 * (2 - 1)+\margin}:\radius) arc ({360/3 * (2 - 1)+\margin}:{360/3 * (2)-\margin}:\radius);
    
        \draw[<-, >=latex, postaction={decorate,decoration={text effects along path,
        text={\ Collect information \ }, text align=center, 
        text effects/.cd, 
          text along path, 
          every character/.style={fill=white, yshift=-0.5ex}}}] ({360/3 * (3 - 1)+\margin}:\radius) arc ({360/3 * (3 - 1)+\margin}:{360/3 * (3)-\margin}:\radius);
          
        \def \radius {1cm}
        \def \margin {30} 
    
        \begin{scope}[shift={(-0.8,1.3)}]
    
        \node (4) [] at ({360/3 * (1 - 1)}:\radius)  {\Huge \color{gray}{\faGlobe}};
        \node (5) [] at ({360/3 * (2 - 1)}:\radius) {};
        \node (6) [circle, minimum size=0.5cm, text centered, draw=gray] at ({360/3 * (3 - 1)}:\radius) {\parbox{0.4cm}{ \tiny \centering \color{gray}{
        $\Delta M$ \\
        $\Delta I$ }}};
        
        \node (7) [] at (0,0) {\parbox{1.8cm}{ \scriptsize \centering \color{gray}{
        Simulate\\
        and \\
        plan }}};

        \draw[<-, >=latex, draw=gray, dashed] ({360/3 * (1 - 1)+\margin}:\radius) arc ({360/3 * (1 - 1)+\margin}:{360/3 * (1)-\margin}:\radius);
        \draw[<-, >=latex, draw=gray, dashed] ({360/3 * (2 - 1)+\margin}:\radius) arc ({360/3 * (2 - 1)+\margin}:{360/3 * (2)-\margin}:\radius);
        \draw[<-, >=latex, draw=gray, dashed] ({360/3 * (3 - 1)+\margin}:\radius) arc ({360/3 * (3 - 1)+\margin}:{360/3 * (3)-\margin}:\radius);
    
        \def \radius {1.4cm}
    
        \end{scope}
    \end{tikzpicture}

\caption{A PDT as a mental model of an agent taking actions in the real world. As new experience is gained, the PDT may be updated by 
changing the structural assumptions $M$ that defined the probability measure $P$, or updating belief with respect to epistemic events through conditioning on 
the new set of information $I$. The changes in structural assumptions and epistemic information are represented by $\Delta M$ and $\Delta I$ respectively.
As part of the planning process, the PDT may simulate possible scenarios as indicated by the inner circle.}
\label{fig:pdt_mental_model}
\end{figure}
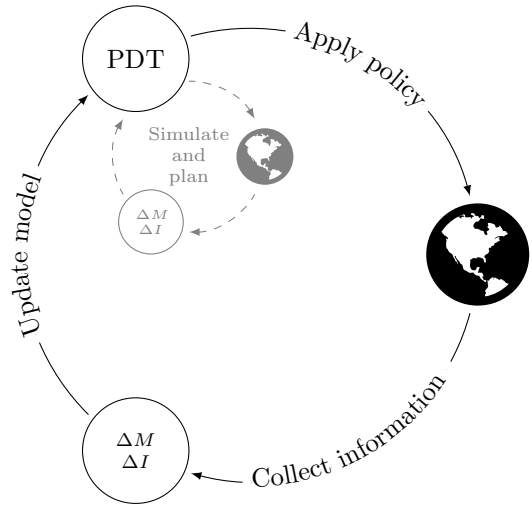

In this kind of scenario, we want to use the PDT to determine an \emph{action} or \emph{policy} for how to act optimally (with respect to some case-specific criterion). 
By a \emph{policy} here we mean the instructions for how to select among multiple actions given the information available at each discrete time step.
We describe this in more detail in \secref{sec:PDT_planning} where we discuss how the PDT is used for \emph{planning}.
When we make use of the PDT in this way, we consider the PDT as a "mental model" of the real physical system, which an agent uses 
to evaluate the potential consequences of actions. The agent then decides on some action to make, observes the outcome, and updates her beliefs about 
the true system, as illustrated in \figref{fig:pdt_mental_model}.

Whether the agent applies a policy or just an action (the first in the policy) before collecting information and updating the probabilistic model depends on 
the type of application at hand. In general it is better to update the model as often as possible, preferably between each action, but the actual computational time needed to perform this updating might make it impossible to achieve in practice. 

\subsection{Mathematical framework of sequential decision making}
\label{sec: math_seq_decisions}

In this section, we briefly recap the mathematical framework of stochastic, sequential decision making in discrete time. We first recall the general framework, and in the following Section \ref{sec: seq_PDT}, we show how this relates to our definition of a PDT.

Let $t = 0, 1, 2, \ldots, N-1$ and consider a discrete time system where the state of the system, $\{x_t\}_{t \geq 1}$, is given by
\begin{equation}
    \label{eq: state_constraint}
    x_{t+1} = f_t(x_t, u_t, w_t), \quad t = 0, 1, 2, \ldots, N-1.
\end{equation}

Here, $x_t$ is the \emph{state of the system} at time $t$, $u_t$ is a \emph{control} and $w_t$ is a noise, or random parameter at time $t$. Note that the control, $u_t$, is a decision which can be made by an \emph{agent} (\emph{the controller}) at time $t$. This control is to be chosen from a set of admissible controls $\mathcal{A}_t$ (possibly, but not necessarily depending on time). Also, $f_t$, $t=0, 1, 2, \ldots, N-1$ are functions mapping from the space of state variables (state space), controls and noise into the set of possible states of $\{x_t\}_{t \geq 0}$. The precise structure of the state space, set of admissible controls and the random parameter space depends on the particular problem under consideration. Note that due to the randomness in $w_t$, $t=0,1,2, \ldots, N-1$, the system state $x_t$ and control $u_t$, $t=1, 2, \ldots, N-1$ also become random variables.

We remark that because of this, the state equation is sometimes written in the following form,
\begin{equation}
    \label{eq: state_eq2}
    x_{t+1}(\omega) = f_t(x_t(\omega), u_t(\omega), \omega) 
\end{equation}
\noindent where $\omega \in \Omega$ is a scenario in a scenario space $\Omega$ (representing the randomness). Sometimes, the randomness is suppressed for notational convenience, so the state equation becomes $x_{t+1} = f_t(x_t, u_t)$, $t = 0, 1, 2, \ldots, N-1$.

Note that in the state equation \eqref{eq: state_constraint} (alternatively, equation \eqref{eq: state_eq2}), $x_{t+1}$ only depends on the previous time step, i.e., $x_{t}, u_{t}, w_{t}$. This is the \emph{Markov property} (as long as we assume that the distribution of $w_t$ does not depend on past values of $w_s$, $s=0,1, \ldots t-1$, only $x_t, u_t$). That is, the next system state only depends on the previous one. Since this Markovian framework is what will be used throughout this paper as we move on to reinforcement learning for a probabilistic digital twin, we focus on this. However, we remark that there is a large theory of sequential decision making which is not based on this Markovianity. This theory is based around maximum principles instead of dynamic programming, see the following Section \ref{sec: DPPvsMP} for more on this.

The aim of the agent is to minimize a cost function under the state constraint \eqref{eq: state_constraint} (or alternatively, \eqref{eq: state_eq2}). We assume that this cost function is of the following, additive form,
\begin{equation}
\label{eq: cost_func}
E\big[g(x_N) + \sum_{t=0}^{N-1} h_t(x_t, u_t, w_t)\big]
\end{equation}
\noindent where the expectation is taken with respect to an a priori given probability measure.
That is, we sum over all \emph{instantaneous rewards} $h_t(x_t, u_t, w_t)$, $t=0, 1, \ldots, N-1$ which depend on the state of the system, the control and the randomness and add a \emph{terminal reward} $g(x_N)$ which only depends on the system state at \emph{the terminal time} $t=N$. This function is called the \emph{objective function}.

Hence, the stochastic sequential decision making problem of the agent is to choose admissible controls $u_t$, $t =0, 1, 2, \ldots, N-1$ in order to,
\begin{equation}
    \label{eq: opt-control}
   \min_{u_t \in \mathcal{A}_t, t \geq 0} \quad  E\big[g(x_N)+\sum_{t=0}^{N-1} h_t(x_t, u_t, w_t) \big]
\end{equation}
\noindent such that
\begin{equation*}
    x_{t+1} = f_t(x_t, u_t, w_t), \quad t = 0, 1, 2, \ldots, N-1.   
\end{equation*}

Typically, we assume that the agent has \emph{full information} in the sense that they can choose the control at time $t$ based on (fully) observing the state process up until this time, but that they are not able to use any more information than this (future information, such as inside information).

This problem formulation is very similar to that of continuous time stochastic optimal control problem.

\begin{remark}(A note on continuous time)
This framework is parallel to that of stochastic optimal control in continuous time. The main differences in the framework in the continuous time case is that the state equation is typically a stochastic differential equation, and the sum is replaced by an integral in the objective function. For a detailed introduction to continuous time control, see e.g., {\O}ksendal \cite{Oksendal}.

\end{remark}

Other versions of sequential decision making problems include inside information optimal control, partial information optimal control, infinite time horizon optimal control and control with various delay and memory effects. One can also consider problems where further constraints, either on the control or the state, is added to problem \eqref{eq: opt-control}.

In Bertsekas \cite{Bertsekas}, the sequential decision making problem \eqref{eq: opt-control} is studied via \emph{the dynamic programming algorithm}. This algorithm is based on \emph{the Bellman optimality principle}, which says that an optimal policy chosen at some initial time, must be optimal when the problem is re-solved at a later stage given the state resulting from the initial choice. 

\subsection{Sequential decision making in the PDT}
\label{sec: seq_PDT}

Now, we show how the sequential decision making framework from the previous section can be used to solve sequential decision making problems in the PDT.

We may apply this sequential decision making framework to our PDT by letting
\[
x_t := \X_t.
\]
That is, the state process for the PDT sequential decision making problem is the random vector of attributes $\X_t$. Note that in Definition \ref{def: PDT}, there is no time-dependency in the attributes $\X$. However, since we are interested in considering sequential decision making in the PDT, we need to assume that there is some sort of development over time (or some indexed set, e.g. information) of the PDT.

Hence, the stochastic sequential decision making problem of the PDT-agent is to choose admissible controls $u_t$, $t =0, 1, 2, \ldots, N-1$ in order to,
\begin{equation}
    \label{eq: opt-control_PDT}
   \min_{u_t \in \mathcal{A}_t, t \geq 0} \quad  E\big[g(\X_N) + \sum_{t=0}^{N-1} h_t(\X_t, u_t, w_t)\big]
\end{equation}
\noindent such that
\[   
 \X_{t+1} = f_t(\X_t, u_t, w_t), \quad t = 0, 1, 2, \ldots, N-1.
\]
Here, the set of admissible controls, $\{u_t\}_{t \geq 0} \in \mathcal{A}$, are problem specific. So are the functions $h_t, g_t$ and $f_t$ for $t \geq 0$. 
Given a particular problem, these functions are defined based on the goal of the PDT-agent as well as the updating of the PDT given new input.

\subsection{Two solution methods for sequential decision making}
\label{sec: DPPvsMP}

In the literature on (discrete time) stochastic sequential decision making, there are two main approaches:

\begin{itemize}
    \item The dynamic programming principle (DPP).
    \item The Pontyagrin maximum principle (MP).
\end{itemize}
The continuous time analogues are the Hamilton-Jacobi-Bellman equations (a particular kind of partial differential equation) and the stochastic maximum principle, respectively.

The DPP is based on the Bellman optimality principle, and the resulting Bellman equation which can be derived from this. Dynamic programming has a few important advantages. It is tractable from an algorithmic perspective because of the backpropagation algorithm naturally resulting from the Bellman equation. Furthermore, the method is always well defined since it is based on working with the value function
(the function that maps states to the optimal value given by \eqnref{eq: opt-control_PDT}, assuming that we start from the given state). 
However, there are some downsides to the DPP method as well. Firstly, DPP requires a Markovian framework (or that we can transform the problem to a Markovian framework). Also, the DPP requires that the Bellman equation holds. This may not be the case if we have problems with for example non-exponential discounting (with respect to time). In this case, we say that there are problems with time-inconsistency, see e.g., Rudloff et al. \cite{Rudloff}. For instance, traditional risk measures such as value-at-risk (VaR) and conditional-value at risk (CVaR) are time-inconsistent in this sense, see Cheridito and Stadje \cite{Cheridito}, and Artzner et al. \cite{Artzner} respectively. Hence, we run into time-inconsistency issues when e.g., minimizing the conditional-value-at-risk of some financial position, if we are using the DPP method. Finally, we cannot have state constraints when using the DPP method, since this causes discontinuity of the value function, see Hao and Li \cite{li2018optimal}.

The alternative approach to solving stochastic sequential decision making problems is via the Pontryagin maximum principle. This method does not require Markovianity or depend on Bellman equation. Hence, there are no problems with time-inconsistency. However, the MP approach is less tractable from an algorithmic point of view. Furthermore, the MP approach requires existence of a minimizing (or maximizing) control. This may not be the case, since it is possible that only limiting control processes converging to the minimum (maximum) exist.

The pros and cons of dynamic programming and the maximum principle approach carry over in continuous time.

From a computational point of view, the dynamic programming method suffers from the curse of dimensionality. When doing numerical backward induction in the DPP, the objective function must be computed for each combination of values. This makes the method too computationally demanding to be applicable in practice for problems where the state space is large, see Agrell and Dahl \cite{Agrell:2020:DOE_SRA} for a discussion of this. Until recently, numerical algorithms based on the maximum principle were not frequently studied in the literature, an exception is Bonnans \cite{Bonnans}. However, the MP approach leads to systems of backward differential equations, in the continuous case, which are often computationally demanding and also less tractable from an algorithmic point of view than the DPP method. However, with the advances of machine learning over the past decade, some new approaches based on the MP approach using deep learning have been introduced, see Li et al. \cite{li2017maximum}. 

Actually, reinforcement learning (RL) is essentially the DPP method. Hence, RL algorithms also suffer from the curse of dimensionality, see Sutton and Barto \cite{sutton1998introduction}. This means that most RL algorithms become less efficient when the dimension of the state space increases. However, by using function approximation the curse of dimensionality can often be efficiently handled, see Arulkumaran et al. \cite{arulkumaran2017brief}. 

The purpose of this paper is to build on this literature by connecting deep reinforcement learning (so essentially, the dynamic programming method) to probabilistic digital twins in order to do planning with respect to the PDT. This is the topic of the following section.

\subsection{Planning in the PDT}
\label{sec:PDT_planning}
In this section, we discuss how the PDT can be used for \emph{planning}. That is, how we use the PDT to identify an optimal policy, without acting in the real world, but by instead simulating what will happen in the real world given that the agent chooses specific actions (or controls, as they are called in the sequential decision making literature, see Section \ref{sec: math_seq_decisions}). We use the PDT as a tool to find a plan (policy), or a single action (first action of policy), to perform in the real world.

In order to solve our sequential decision making problem in the PDT, we have chosen to use a reinforcement learning formulation. As remarked in Section \ref{sec: DPPvsMP}, this essentially corresponds to choosing the dynamic programming method for solving the optimal control problem (as opposed to a maximum principle approach). Because we will use a DPP approach, we need all the assumptions that come with this, see the discussion in Section \ref{sec: DPPvsMP}: A Markovian framework, or the possibility of transforming the problem to something Markovian. We need the Bellman equation to hold in order to avoid issues with time-inconsistency. In order to ensure this, we for example need to use exponential discounting and not have e.g., conditional expectation of state process in a non-linear way in the objective function. Finally, our planning problem cannot have state constraints.

\begin{remark}
Instead of using the DPP to solve the planning problem, we could use a maximum principle approach. One possible way of doing this in practice, is by using one of the MP based algorithms found in Li et al. \cite{li2017maximum}, instead of using reinforcement learning. By this alternative approach, we avoid the Markovianity requirement, possible time-inconsistency issues and can allow for state constraints (via a Lagrange multiplier method - see e.g., Dahl and Stokkereit \cite{dahl2016stochastic}). This topic is beyond the scope of this paper, but is a current work in progress. 
\end{remark}

Starting with an initial PDT as a digital representation of a physical system given our current knowledge, we assume that there are two ways to update the PDT: 

\begin{enumerate}
    \item Changing or updating the structural assumptions $M$, and hence the probability measure $P_M$.
    \item Updating the information $I$.
\end{enumerate}

The structural assumptions $M$ are related to the probabilistic model for $\X$. Recall from Section \ref{sec:epistemic_conditioning}, that these assumptions define the probability measure $P_M$. Often, this probability measure is taken as given in stochastic modeling. However, in practice, probability measures are not given to us, but \emph{decided} by analysts based on previous knowledge. Hence, the structural assumptions $M$ may be updated because of new knowledge, external to the model, or for other reasons the analysts view as important.

Updating the information is our main concern in this paper, since this is related to the agent making costly decisions in order to gather more information. An update of the information also means (potentially) reducing the epistemic uncertainty in the PDT. Optimal information gathering in the PDT will be discussed in detail in the following Section \ref{sec:optimal_info_gathering}.

\subsection{MDP, POMDP and its relation to DPP}
\label{sec: MDP_POMDP_DPP}

In this section, we briefly recall the definitions of Markov decision processes, partially observable Markov decision processes and explain how these relate to the seuqntial decision making framework of Section \ref{sec: math_seq_decisions}.

\emph{Markov decision processes} (MDP) are discrete-time stochastic control processes of a specific form. An MDP is a tuple 
\[
(S, A, P_a, R_a),
\] 
\noindent where $S$ is a set of states (the state space) and $A$ is a set of actions (action space). Also,
\[
P_a(s, s') = P_a(s_{t+1} = s' \mid a_t=a, s_t=s)
\]
\noindent is the probability of going from state $s$ at time $t$ to state $s'$ at time $t+1$ if we do action $a$ at time $t$. Finally, $R_a(s,s')$ is the instantaneous reward of transitioning from state $s$ at time $t$ to state $s'$ at time $t+1$ by doing action $a$ (at time $t$).

An MDP satisfies the \emph{Markov property}, so given that the process is in state $s$ and will be doing $a$ at time $t$, the next state $s_{t+1}$ is conditionally independent of all other previous states and actions. 

\begin{remark}\label{remark: MDP_DPP}(MDP and DPP)\\
Note that this definition of an MDP is essentially the same as our DPP framework of Section \ref{sec: math_seq_decisions}. In the MDP notation, we say \emph{actions}, while in the control notation, it is common to use the word \emph{control}. In Section \ref{sec: math_seq_decisions}, we talked about \emph{instantaneous cost functions}, but here we talk about \emph{instantaneous rewards}. Since minimization and maximization problems are equivalent (since $\inf\{\cdot \}= - \sup \{-\cdot\}$), so are these two concepts. Furthermore, the definition of the transition probabilities $P_a$ in the MDP framework corresponding to the Markov assumption of the DPP method. In both frameworks, we talk about the system states, though in the DPP framework we model this directly via equation \eqref{eq: state_constraint}.
\end{remark}

A generalization of MDP are \emph{partially observable Markov decision processes} (POMDPs). While an MDP is a $4$-tuple, a POMDP is a $6$-tuple,

\[
(S, A, P_a, R_a, \bar{\Omega}, O).
\]
Here (like before), $S$ is the state space, $A$ is the action space, $P_a$ give the conditional transition probabilities between the different states in $S$ and $R_a$ give the instantaneous rewards of the transitions for a particular action $a$. 

In addition, we have $\bar{\Omega}$, which is a \emph{set of observations}. In contrast to the MDP framework, with POMDP, the agent no longer observes the state $s$ directly, but only an observation $o \in \bar{\Omega}$. Furthermore, the agent knows $O$ which is a \emph{set of conditional observation probabilities}. That is,

\[
O(o \mid s', a)
\]
\noindent is the probability of observing $o \in \bar{\Omega}$ given that we do action $a$ from state $s'$. 

The objective of the agent in the POMDP sequential decision problem is to choose a policy, that is actions at each time, in order to 
\begin{equation}
\label{eq: infinite}
    \max_{\{a_t\} \in A} E\big[ \sum_{t=0}^{T} \lambda^t r_t \big]
\end{equation}
\noindent where $r_t$ is the reward earned at time $t$ (depending on $s_t, a_t$ and $s_{t+1}$),
and $\lambda \in [0, 1]$ is a number called the discount factor. The discount factor can be used to introduce a preference for immediate rewards as opposed to more 
distant rewards, which may be relevant for the problem at hand, or used just for numerical efficiency. 
Hence, the agent aims to maximize their expected discounted reward over all future times. Note that is it also possible to consider problem \eqref{eq: infinite} over an infinite time horizon or with a separate terminal reward function as well. This is similar to the DPP sequential decision making framework of Section \ref{sec: math_seq_decisions}.

In order to solve a POMDP, it is necessary to include memory of past actions and observations. Actually, the inclusion of partial observations means that the problem is no longer Markovian. However, there is a way to Markovianize the POMDP by transforming the POMDP into a \emph{belief-state MDP}. In this case, the agent summarizes all information about the past in a belief vector $b(t)$, which is updated as time passes. See \cite{wiering2012reinforcement}, Chapter 12.2.3 for details. 

\subsection{MDP (and POMDP) in the PDT framework}
\label{sec:POMDP_PDT}
In this section, we show how the probabilistic digital twin can be incorporated in a reinforcement learning framework, in order to solve sequential decision problems in the PDT.

In Section \ref{sec: seq_PDT}, we showed how we can use the mathematical framework of sequential decision making to solve optimal control problems for a PDT-agent. Also, in Section \ref{sec: MDP_POMDP_DPP}, we saw (in Remark \ref{remark: MDP_DPP}) that the MDP (or POMDP in general) framework essentially corresponds to that of the DPP. In theory, we could use the sequential decision making framework and the DPP to solve optimal control problems in the PDT. However, due to the curse of dimensionality, this will typically not be practically tractable (see Section \ref{sec: DPPvsMP}). In order to resolve this, we cast the PDT sequential decision making problem into a reinforcement learning, in particular a MDP, framework. This will enable us to solve the PDT optimal control problem via deep reinforcement learning, in which there are suitable tools to overcome the curse of dimensionality.

To define a decision making process in the PDT as a MDP, we need to determine our state space, action space, (Markovian) transition probabilities and a reward function. 

\begin{itemize}
    \item The action space $A$: These are the possible actions within the PDT. These may depend on the problem at hand. In the next Section \ref{sec:optimal_info_gathering}, we will discuss optimal information gathering, where the agent can choose between different types of experiments, at different costs, in order to gain more information. In this case, the action space is the set of possible decisions that the agent can choose between in order to attain more information. 
    
    \item The state space $S$: 
    We define a state as a PDT (or equivalently a \emph{version} of a PDT that evolves in discrete time $t = 0, 1, \dots$). 
    A PDT represents our belief about the current \emph{physical} state of a system, 
    and it is defined by some initial assumptions together with the information acquired through time. 
    In practice, if the structural assumptions are not changed, we may let the information available at the current time represent a state.
    
    This means that our MDP will consist of \emph{belief-states}, represented by information, from which inference about the true physical state can be made.
    This is a standard way of creating a MDP from a POMDP, so we can view the \emph{PDT state-space} as a space of \emph{beliefs} about 
    some underlying partially observable physical state.
    
    Starting from a PDT, we define the state space as all updated PDTs we can reach by taking actions in the action space $A$. 
    
    \item The transition probabilities $P_a$: Based on our chosen definition of the state space, the transition probabilities are the probabilities of going from one level of information to another, given the action chosen by the agent. For example, if the agent chooses to make decision (action) $d$, what is the probability of going from the current level of information to another (equal or better) level. This is given by epistemic conditioning of the PDT with respect to the given information set $I = \{(d,o)\}$ based on the decisions $d$ the new observation $o$. When it comes to updates of the structural assumptions $M$, we consider this as deterministic transitions. 
    
    \item The reward $R_a$: The reward function, or equivalently, cost function, will depend on the specific problem at hand. To each action $a \in A$, we assume that we have an associated reward $R_a$. In the numerical examples in Section \ref{sec: deep_RL}, we give specific examples of how these rewards can be defined.
\end{itemize}

As mentioned in Section \ref{sec:PDT_planning}, there are two ways to update the PDT: Updating the structural assumptions $M$ and updating the information $I$. If we update the PDT by (only) adding to the information set $I$, we always have the Markov property.  

If we also update $M$, then the preservation of the Markov property is not given. In this case, using a maximum principle deep learning algorithm instead of the DPP based deep RL is a possibility, see \cite{li2017maximum}. 

\begin{remark}
Note that in the case where we have a very simple PDT with only discrete variables and only a few actions, then the RL approach is not necessary. In this case, the DPP method as done in traditional optimal control works well, and we can apply a planning algorithm to the PDT in order to derive an optimal policy. However, in general, the state-action space of the PDT will be too large for this. Hence, traditional planning algorithms, and even regular RL may not be feasible due to the curse of dimensionality.
In this paper, we will consider deep reinforcement learning as an approach to deal with this. We discuss this further in Section \ref{sec: deep_RL}. 
\end{remark}

Note that what determines an optimal action or policy will of course depend on what \emph{objective} the outcomes are measured against. 
That is, what do we want to achieve in the real world? There are many different objectives we could consider. 
In the following we present one generic objective related to optimal information gathering, where the PDT framework is suitable.

\subsection{Optimal information gathering}
\label{sec:optimal_info_gathering}
A generic, but relevant, objective in optimal sequential decision making is simply to "improve itself". That is, to reduce epistemic uncertainty with respect to some quantity of interest. 
Another option, is to consider maximizing the Kullback-Leibler divergence with respect to epistemic uncertainty as a general objective. 
This would mean that we aim to collect the information that "will surprise us the most".

By definition, a PDT contains an observational model related to the data generating process (the epistemic conditioning relies on this).  
This means that we can simulate the effect of gathering information, 
and we can study how to do this optimally. In order to define what we mean by an optimal strategy for gathering information, we then have to specify the following,

\begin{itemize}
    \item \emph{Objective:} What we need the information for. For example, what kind of decision do we intend to support using the PDT? 
    Is it something we want to estimate? What is the required accuracy needed? For instance, we might want to reduce epistemic uncertainty with 
    respect to some quantity, e.g., a risk metric such as a failure probability, expected extreme values etc. 
    \item \emph{Cost:} The cost related to the relevant information-gathering activities. 
\end{itemize}

Then, from the \emph{PDT} together with a specified \emph{objective} and \emph{cost}, one alternative is to define the optimal strategy as the strategy that minimizes the (discounted) expected cost needed to achieve the objective (or equivalently achieves the objective while maximizing reward).

\begin{ex}(Coin flip -- information gathering)
\label{ex:coin_inf_gathering}
    Continuing from Example \ref{ex: coin_robust}, imagine that before making your final bet, you can flip the coin as many times as you like in order to learn about $\theta$.
    Each of these test flips will cost $10.000 \ \$$. You also get the opportunity to replace the coin with a new one, at the cost of $100.000 \ \$$.
    
    An interesting problem is now how to select an optimal strategy for when to \emph{test}, \emph{bet} or \emph{replace} in this game. 
    And will such a strategy be robust? What if there is a limit on the total number of actions than can be performed? 
    In \secref{sec:coin_example_RL} we illustrate how reinforcement learning can be applied to study this problem, 
    where the coin represents a component with reliability $\theta$, that we may \emph{test}, \emph{use} or \emph{replace}.
\end{ex}

\section{Deep Reinforcement Learning with PDTs}
\label{sec: deep_RL}
In this section we give an example of how reinforcement learning can be used for planning, i.e. finding an optimal action or policy, with a PDT.
The reinforcement learning paradigm is especially relevant for problems where the state and/or action space is large, or dynamical models where 
specific transition probabilities are not easily attainable but where efficient sampling is still feasible. In probabilistic modelling of complex physical phenomena, 
we often find ourselves in this kind of setting. 

\subsection{Reinforcement Learning (RL)}
Reinforcement learning, in short, aims to optimize sequential decision problems through sampling from a MDP (Sutton and Barto \cite{sutton1998introduction}). 
We think of this as an agent taking actions within an environment, following some policy $\pi(a|s)$, which gives the probability of taking action $a$ if the agent 
is currently at state $s$. Generally, $\pi(a|s)$ represents a (possibly degenerate) probability distribution over actions $a \in A$ for each $s \in S$. 
The agent's objective is to maximize the amount of reward it receives over time, and a policy $\pi$ that achieves this is called an \emph{optimal policy}.

Given a policy $\pi$ we can define the \emph{value} of a state $s \in S$ as
\begin{equation}
    \label{eq:RL_value}
    v_{\pi}(s) = E\big[ \sum_{t=0}^{T} \lambda^t r_t \mid s_0 = s \big]
\end{equation}
where $r_t$ is the reward earned at time $t$ (depending on $s_t, a_t$ and $s_{t+1}$), given that the agent follows policy $\pi$ starting from $s_0 = s$. That is, 
for $P_a$ and $R_a$ given by the MDP, $a_t \sim \pi(a_t | s_t)$, $s_{t+1} \sim P_{a_t}(s_t, s_{t+1})$ and $r_t \sim R_{a_t}(s_t, s_{t+1})$.
Here we make use of a discount factor $\lambda \in [0, 1]$ in the definition of cumulative reward. 
If we want to consider $T = \infty$ (continuing tasks) instead of $T < \infty$ (episodic task), then $\lambda < 1$ is generally necessary. 

The \emph{optimal value function} is defined as the one that maximises \eqnref{eq:RL_value} over all policies $\pi$. 
The optimal action at each state $s \in S$ then corresponds to acting greedily with respect to this value function, i.e. 
selecting the action $a_t$ that in expectation maximises the value of $s_{t+1}$. 
Likewise, it is common to define the \emph{action-value} function $q_{\pi}(s, a)$, which corresponds to the expected cumulative return 
of first taking action $a$ in state $s$ and following $\pi$ thereafter. 
RL generally involves some form of Monte Carlo simulation, where a large number of \emph{episodes} are sampled from the MDP, 
with the goal of estimating or approximating the optimal value of sates, state-action pairs, or an optimal policy directly. 

Theoretically this is essentially equivalent to the DPP framework, 
but with RL we are mostly concerned with problems where optimal solutions cannot be found and some form of approximation is needed. 
By the use of flexible approximation methods combined with adaptive sampling strategies, RL makes it possible to 
deal with large and complex state- and action spaces.

\subsection{Function approximation}
One way of using function approximation in RL is to define a parametric function $\hat{v}(s, \w) \approx v_{\pi}(s)$, 
given by a set of weights $\w \in \mathbb{R}^d$, and try to learn the value function of an optimal policy by finding an appropriate value for $\w$.
Alternatively, we could approximate the value of a state-action pair, $\hat{q}(s, a, \w) \approx q_{\pi}(s, a)$, or a policy $\hat{\pi}(a | s, \w) \approx \pi(a | s)$.
The general goal is then to optimize $\w$, using data generated by sampling from the MDP, and the RL literature contains many different algorithms designed for this purpose. 
In the case where a neural network is used for function approximation, it is often referred to as \emph{deep reinforcement learning}. 
One alternative, which we will make use of in an example later on, is the deep Q-learning (DQN) approach as introduced by van Hasselt et al. \citep{Hasselt:2016:DQN}, which represents the value of a set of $m$ actions 
at a state $s$ using a multi-layered neural network
\begin{equation}
    \hat{q}(s, \w) : S \rightarrow \mathbb{R}^m.
\end{equation}
Note here that $\hat{q}(s, \w)$ is a function defined on the state space $S$. In general, any approximation of the value functions $v$ or $q$, or the policy $\pi$ are defined 
on $S$ or $S \times A$. A question that then arises, is how can we define parametric functions on the state space $S$ when we are dealing with PDTs? 
We can assume that we have control over the set of admissible actions $A$, in the sense that this is something we define, 
and creating parametric functions defined on $A$ should not be a problem. But as discussed in \secref{sec:POMDP_PDT}, 
$S$ will consist of \emph{belief}-states.

\subsection{Defining the state space}
We are interested in an MDP where the transition probabilities $P_a(s, s')$ corresponds to updating a PDT as a consequence of action $a$. 
In that sense, $s$ and $s'$ are PDTs. 
Given a well-defined set of admissible actions, the state space $S$ is then the set of all PDTs that can be obtained starting from some initial state $s_0$, within some defined 
horizon. 

Recall that going from $s$ to $s'$ then means keeping track of any changes made to the structural assumptions $M$ and the information $I$, as illustrated in \figref{fig:pdt_mental_model}. 
From now on, we will for simplicity assume that updating the PDT only involves epistemic conditioning with respect to the information $I$. 
This is a rather generic situation. Also, finding a way to represent changes in $M$ will have to be handled for the specific use case under consideration. 
Assuming some initial PDT $s_0$ is given, any state $s_t$ at a later time $t$ is then uniquely defined by the set of information $I_t$ available at time $t$. 
Representing states by information in this way is something that is often done to transform a POMDP to a MDP. 
That is, although the true state $s_t$ at time $t$ is unknown in a POMDP, the information $I_t$, and consequently our \emph{belief} about $s_t$, is always know at time $t$.  
Inspired by the POMDP terminology, we may therefore view a PDT as a belief-state, which seems natural as 
the PDT is essentially a way to encode our beliefs about some real physical system. 

Hence, we will proceed with defining the state space $S$ as the \emph{information state-space}, which is the set of all sets of information $I$.
Although this is a very generic approach, we will show that there is a way of defining a flexible parametric class of functions on $S$. 
But we must emphasize that that if there are other ways of compressing the information $I$, for instance due to conjugacy in the epistemic updating, 
then this is probably much more efficient. \exref{ex: coin_state_space} below shows exactly what we mean by this.

\begin{ex}(Coin flip -- information state-space)
\label{ex: coin_state_space}
    In the coin flip example (Example \ref{ex:coin_1}), all of our belief with respect to epistemic uncertainty is represented by the 
    number $\psi = P(\theta = \theta_1)$. Given some observation $Y = y \in \{0, 1 \}$, the epistemic conditioning corresponds to
    \begin{equation*}
        \psi \rightarrow \frac{\beta_1(y) \psi}{\beta_1(y) \psi + \beta_2(y) (1 - \psi)},
    \end{equation*}
    where, for $j = 1, 2$, $\beta_j(y) = \theta_j$ if $y = 0$ and $\beta_j(y) = 1-\theta_j$ if $y = 1$.
    
    In this example, the information state-space consists of all sets of the form $I_t = \{ y_1, \dots, y_t \}$ where each $y_i$ is binary.
    However, if the goal is to let $I_t$ be the representation of a PDT, we could just as well use $\psi_t$, 
    i.e. define $S = [0, 1]$ as the state space. Alternatively, the number of heads and tails (0s and 1s) provides the same information, 
    so we could also make use of $S = \{ 0, \dots, N \} \times \{ 0, \dots, N \}$ where $N$ is an upper limit on the total number of flips we consider.
\end{ex}

\subsection{Deep learning on the information state-space}
Let $S$ be a set of sets $I \subset \mathbb{R}^d$. We will assume that each set $I \in S$ consists of a finite number of elements $y \in \mathbb{R}^d$, 
but we do not require that all sets $I$ have the same size. We are interested in functions defined on $S$.

An important property of any function $f$ that takes a set $I$ as input, is permutation invariance. 
I.e. $f(\{ \y_1, \dots, \y_N \}) = f(\{ \y_{\kappa(1)}, \dots, \y_{\kappa(N)} \})$ for any permutation $\kappa$. 
It can been shown that under fairly mild assumptions, that such functions have the following decomposition 
\begin{equation}
    \label{eq:sum_decomp}
    f(I) = \rho \left(  \sum_{\y \in I} \phi(\y)  \right).
\end{equation}
These sum decompositions were studied by Zaheer et al. \citep{Zaheer:2017:DeepSets} and later by Wagstaff et al. \citep{Wagstaff:2019:limitations}, which showed that if 
$|I| \leq p$ for all $I \in S$, then any continuous function $f : S \rightarrow \mathbb{R}$ can be written as \eqnref{eq:sum_decomp} for some suitable functions
$\phi : \mathbb{R}^d \rightarrow \mathbb{R}^p$ and $\rho : \mathbb{R}^p \rightarrow \mathbb{R}$.
The motivation in \citep{Zaheer:2017:DeepSets, Wagstaff:2019:limitations} was to enable supervised learning of permutation invariant and set-valued functions, 
by replacing $\rho$ and $\phi$ with flexible function approximators, such as Gaussian processes or neural networks. 
Other forms of decomposition, by replacing the \emph{summation} in \eqnref{eq:sum_decomp} with something else that can be learned, has also been considered by Soelch et al.  \citep{Soelch:2019:aggregation}.
For reinforcement learning, we will make use of the form \eqnref{eq:sum_decomp} to represent functions defined on the information states space $S$, 
such as $\hat{v}(s, \w)$, $\hat{q}(s, a, \w)$, or $\hat{\pi}(a | s, \w)$, using a neural network with parameter $\w$. 
In the remaining part of this paper we present two examples showing how this works in practice. 

\subsection{The "coin flip" example}
\label{sec:coin_example_RL}
Throughout this paper we have presented a series of small examples involving a biased coin, represented by $\textbf{X} = (Y, \theta)$. In 
\exref{ex:coin_inf_gathering} we ended by introducing a game where the player has to select whether to 
\emph{bet on}, \emph{test} or \emph{replace} the coin. 
As a simple illustration we will show how reinforcement learning can be applied in this setting. 

But now, we will imagine that the coin $Y$ represents a \emph{component}
in some physical system, where $Y = 0$ corresponds to the component functioning and $Y = 1$ represents failure. 
The probability $P(Y = 1) = 1 - \theta$ is then the components \emph{failure probability}, and we say that $\theta$ is the \emph{reliability}.

For simplicity we assume that $\theta \in \{ 0.5, 0.99 \}$, and that our initial belief is $P(\theta = 0.5) = 0.5$. That is, when we buy a new component, there is a 50 \% chance of getting a "bad" component (that fails 50 \% of the time), and consequently a 50 \% probability of getting a "good" component (that fails 1 \% of the time).

We consider a project going over $N = 10$ days. Each day we will decide between one of the following 4 actions:
\begin{enumerate}
    \item Test the component (flip the coin once). Cost $r = -10.000 \$$. 
    \item Replace the component (buy a new coin). Cost $r = -100.000 \$$.
    \item Use the component (bet on the outcome). Obtain a reward of $r = 10^6 \$$ if the component works ($Y = 0$) and a cost of $r = -10^6 \$$ if the component fails ($Y = 1$).
    \item Terminate the project (set $t = N$), $r = 0$.
\end{enumerate}

We will find a deterministic policy $\pi : S \rightarrow A$ that maps from the information state-space to one of the four actions. The information state-space $S$ 
is here represented by the number of days left of the project, $n = N - t$, and the set $I_t$ of observations of the component that is currently in use at time $t$.
If we let $S_Y$ contain all sets of the form $I = \{ Y_1, \dots, Y_t \}$, for $Y_t \in \{ 0, 1 \}$ and $t < N$, then 
\begin{equation}
    \label{eq:coin_statespace_RL}
    S = S_Y \times \{ 1, \dots, N \}
\end{equation}
represents the information state-space. 
In this example we made use of the deep Q-learning (DQN) approach described by van Hasselt et al. \citep{Hasselt:2016:DQN}, where we define a neural network 
\begin{equation*}
    \hat{q}(s, \w) : S \rightarrow \mathbb{R}^4,
\end{equation*} 
that represents the action-value of each of the four actions. The optimal policy is then obtained by at each state $s$ selecting the 
action corresponding to the maximal component of $\hat{q}$.

We start by finding a policy that optimizes the cumulative reward over the $10$ days (without discounting). 
As it turns out, this policy prefers to "gamble" that the component works rather than performing tests. 
In the case where the starting component is reliable (which happens 50 \% of the time), a high reward can be obtained by selection action 3 at every opportunity. 
The general "idea" with this policy, is that if action 3 results in failure, the following action is to replace the component (action 2), unless there are few days left of the project in which case 
action 0 is selected. We call this the "unconstrained" policy. 

\begin{figure}[h]
    \center{\includegraphics[width=0.5\textwidth, trim={0.8cm 0.1cm 0.3cm 1.1cm},clip]{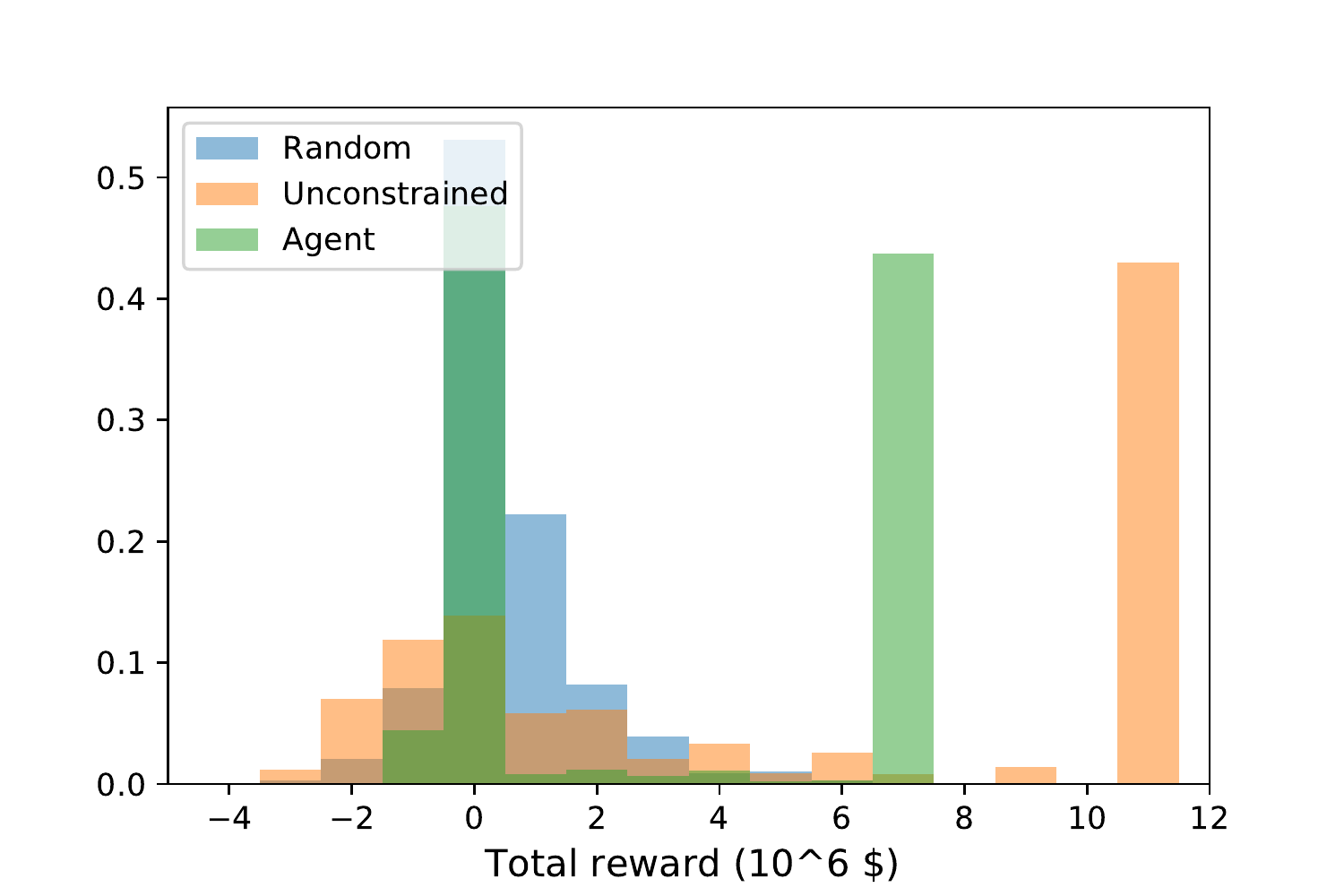}}
    \caption{Total reward after $1000$ episodes for a random policy, the unconstrained policy, and the agent which is subjected to the constraint that action 3 is not allowed 
    unless $P(\theta = 0.99) > 0.9$.}
    \label{fig:RL_coin}
\end{figure}

Although the unconstrained policy givens the largest expected reward, there is an approximately 50 \% chance that it will produce a failure, i.e. that action 3 is selected with 
$Y = 1$ as the resulting outcome. One way to reduce this failure probability, is to introduce  
the constraint that action 3 (using the component) is not allowed unless we have a certain level of confidence 
in that the component is reliable. 
We introduced this type of constraint by requiring that $P(\theta = 0.99) > 0.9$ (a constraint on epistemic uncertainty). 
The optimal policy under this constraint will start with running experiments (action 1), before deciding whether to replace (action 2), use the component (action 3), 
or terminate the project (action 0).
\figref{fig:RL_coin} shows a histogram of the cumulative reward over $1000$ simulated episodes, for the constrained and unconstrained policies obtained by RL, 
together with a completely random policy for comparison.

In this example, the information state-space could also be defined in a simpler way, as explained in \exref{ex: coin_state_space}. 
As a result the reinforcement learning task will be simplified. Using the different state-space representations, we obtained the same results shown in \figref{fig:RL_coin}. 
Finally, we should note that in the case where defining the state space as in \eqnref{eq:coin_statespace_RL} is necessary, 
the constraint $P(\theta = 0.99) > 0.9$ is not practical. That is, if we could estimate this probability efficiently, then we also have 
access to the compressed information state-space. One alternative could then be to instead consider the uncertain failure probability 
$p_f(\theta) = P(Y = 1 \mid \theta )$, and set a limit on e.g. $E[p_f] + 2 \cdot \text{Std}(p_f)$. 
This is the approach taken in the following example concerning failure probability estimation.

\subsection{Corroded pipeline example}
\label{sec:corroded_pipeline_RL}
Here we revisit the corroded pipeline example from Agrell and Dahl \citep{Agrell:2020:DOE_SRA} which we introduced in \secref{sec:corr_pipeline}. 
In this example, we have specified epistemic uncertainty with respect to model discrepancy, the size of a defect, and the capacity $p_{\text{FE}}$ coming from a 
Finite Element simulation. If we let $\theta$ be the epistemic generator, we can write the failure probability conditioned on epistemic information as $p_f(\theta) = P(g \leq 0 \mid \theta)$.
In \citep{Agrell:2020:DOE_SRA} the following objective was considered: Determine with confidence whether $p_f(\theta) < 10^{-3}$. 
That is, when we consider $p_f$ as a purely epistemic random variable, we want to 
either confirm that the failure probability is less than the target $10^{-3}$ (in which case we can continue operations
as normal), or to detect with confidence that the target is exceeded (and we have to intervene). 
Will say the the objective is achieved if we obtain either $E[p_f] + 2 \cdot \text{Std}(p_f) < 10^{-3}$ or $E[p_f] - 2 \cdot \text{Std}(p_f) > 10^{-3}$
(where $E[p_f]$ and $\text{Std}(p_f)$ can be efficiently approximated using the method developed in \citep{Agrell:2020:DOE_SRA}).
There are three ways in which we can reduce epistemic uncertainty:
\begin{enumerate}
    \item \textbf{Defect measurement:} Noise perturbed measurement that reduces uncertainty in the defect size $d$
    \item \textbf{Computer experiment:} Evaluate $p_{\text{FE}}$ at some selected input $(D, t, s, d, l)$, to reduce uncertainty in the 
    surrogate $\hat{p}_{\text{FE}}$ used to approximate $p_{\text{FE}}$.
    \item \textbf{Lab experiment:} Obtain one observation of $X_{\text{m}}$, which reduces uncertainty in $\mu_{\text{m}}$.
\end{enumerate}

The set of information corresponding to defect measurements is $I_{\text{Measure}} \subset \mathbb{R}$ as each measurement is a real valued number. 
Similarly, $I_{\text{Lab}} \subset \mathbb{R}$ as well, and $I_{\text{FE}} \subset \mathbb{R}^6$ when we consider a vector $\y \in \mathbb{R}^6$ as 
an experiment $[D, t, s, d, l, p_{\text{FE}}]$. Actually, in this example we may exploit some conjugacy in in the representation of $I_{\text{Measure}}$ 
and $I_{\text{Lab}}$ as discussed in \exref{ex: coin_state_space} (see \citep{Agrell:2020:DOE_SRA} for details), so we 
can define the information state-space as $S = S_{\text{FE}} \times \mathbb{R}^2$, where $S_{\text{FE}}$ consists of finite subsets of $\mathbb{R}^6$.

We will use RL to determine which of the three types of experiment to perform, and define the action space $A = \{ \text{Measurement}, \text{FE}, \text{Lab} \}$. 
Note that when we decide to run a computer experiment, we also have to specify the input $(D, t, s, d, l)$. 
This is a separate decision making problem regarding design of experiments. For this we make use of the myopic (one-step lookahead) method developed in \citep{Agrell:2020:DOE_SRA}, 
although one could in principle use RL for this as well. 
This kind of decision making, where one first decides between different types of task to perform, and then proceed 
to find a way to perform the selected task optimally, is often referred to as hierarchical RL in the reinforcement learning literature.
Actually, \citep{Agrell:2020:DOE_SRA} considers a myopic alternative for also selecting between the different types of experiments, 
and it was observed that this might be challenging in practice if there are large differences in cost between the experiments. 
This was the motivation for studying the current example, where we now define the reward (cost) $r$ as a direct consequence of $a \in A$ as follows: 
$r = -10$ for $a = \text{Measurement}$, $r = -1$ for $a = \text{Lab}$ and $r = -0.1$ for $a = \text{FE}$.

In this example we also made use of the DQN approach of van Hasselt et al. \citep{Hasselt:2016:DQN}, where we define a neural network 
\begin{equation*}
    \hat{q}(s, \w) : S = S_{\text{FE}} \times \mathbb{R}^2 \rightarrow \mathbb{R}^3,
\end{equation*} 
that gives, for each state $s$, the (near optimal) value of each of the three actions. 
We refrain from describing all details regarding the neural network and the specific RL algorithm, as the main purpose with this example is for illustration. 
But we note that two important innovations in the DQN algorithm, the use of a target network and experience replay as proposed in \citep{Mnih:2015:DQN_nature}, was necessary for this to work.  

The objective in this RL example is to estimate a failure probability using as little resources as possible. 
If an agent achieves the criterion on epistemic uncertainty reduction, that the expected failure probability plus/minus two standard deviations is either above or below the target value, 
we say that the agent has succeeded and we report the sum of the cost of all performed experiments. We also set a maximum limit of $40$ experiments. I.e. after $40$ tries the agent has failed. 
To compare the policy obtained by RL, we consider the random policy that selects between the three actions uniformly at random. 
We also consider a more "human like" benchmark policy, that corresponds to first running $10$ computer experiments, followed by one lab experiment 
then one defect measurement, then $10$ new computer experiments, and so on. 

\begin{figure}[h]
    \center{\includegraphics[width=0.5\textwidth, trim={0.1cm 0.1cm 0.3cm 1.1cm},clip]{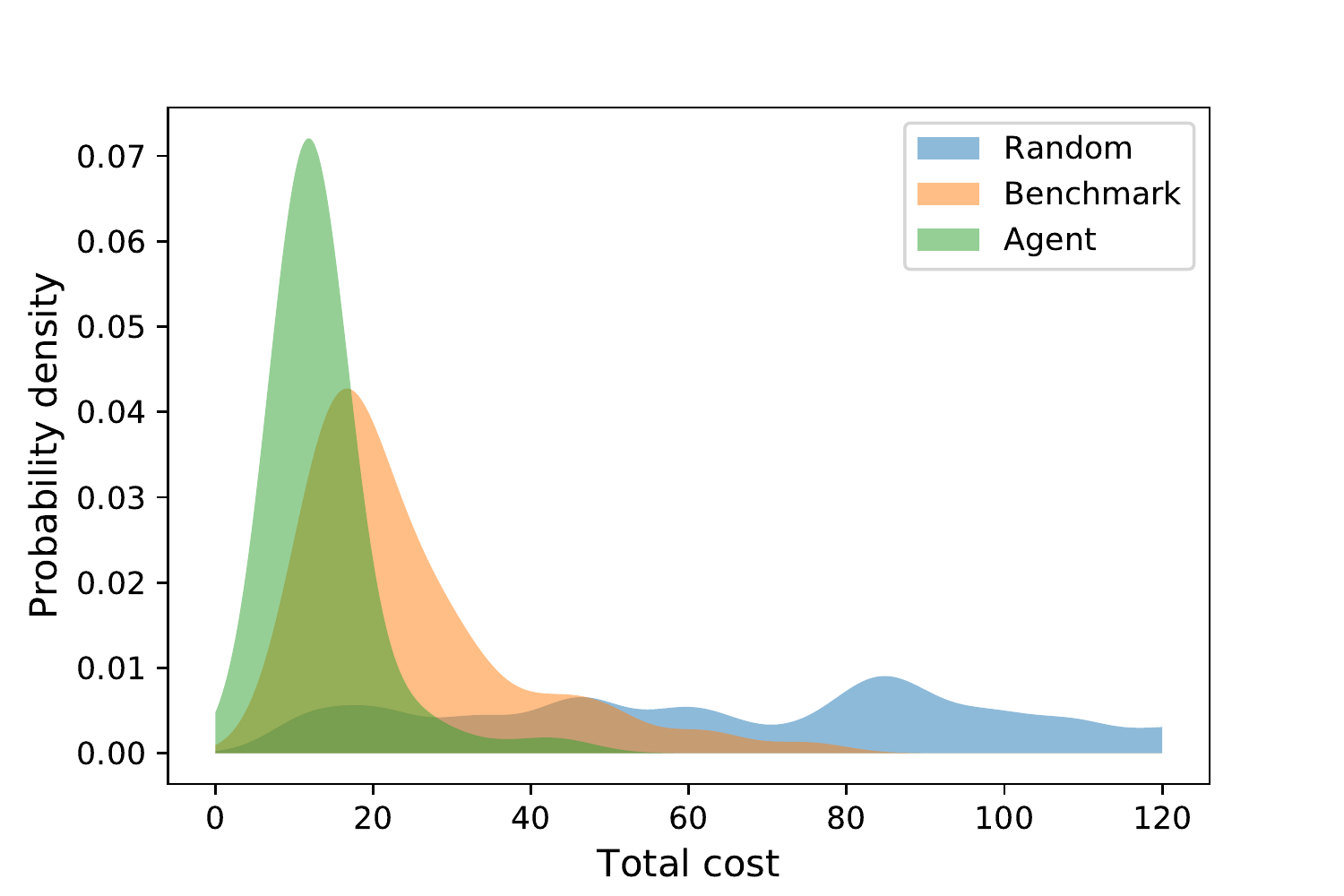}}
    \caption{Total cost (negative reward) after $100$ successful episodes. For the random and benchmark policy, the success rate was around 60\% (to achieve the objective within $40$ experiments in total), whereas 94 \% was successful for the RL agent.}
    \label{fig:RL_pipeline}
\end{figure}

The final results from simulating $100$ episodes with each of the three policies is shown in \figref{fig:RL_pipeline}.

\section{Concluding remarks}
\label{sec: conclusion}
To conclude our discussion, we recall that in this paper, we have:
\begin{itemize}
    \item Given a measure-theoretic discussion of epistemic uncertainty and formally defined epistemic conditioning.
    \item Provided a mathematical definition of a probabilistic digital twin (PDT). 
    \item Connected PDTs with sequential decision making problems, and discussed several solution approaches (maximum principle, dynamic programming, MDP and POMDP).
    \item Argued that using (deep) RL to solve sequential decision making problems in the PDT is a good choice for practical applications today.
   \item For the specific use-case of optimal information gathering, we proposed a generic solution using deep RL on the information state-space.
\end{itemize}

Further research in this direction includes looking at alternative solution methods and RL algorithms in order to handle different PDT frameworks. A possible idea is to use a maximum principle approach instead of a DPP approach (as is done in RL). By using one of the MP based algorithms in \cite{li2017maximum}, we may avoid the Markovianity requirement, possible time-inconsistency issues and can also allow for state constraints. For instance, this is of interest when the objective of the sequential decision making problem in the PDT is to minimize a risk measure such as CVaR or VaR. Both of these risk measures are known to cause time-inconsistency in the Bellman equation, and hence, the DPP (and also RL) cannot be applied in a straightforward manner. This is work in progress.

\section*{Acknowledgements}
This work has been supported by grant 276282 from the Research Council of Norway (RCN)
and DNV Group Research and Development (Christian Agrell and Andreas Hafver). The work has also been supported by grant 29989 from the Research Council of Norway as part of the SCROLLER project (Kristina Rognlien Dahl). 
The main ideas presented in this paper is the culmination of input from recent research activities related to 
risk, uncertainty, machine learning and probabilistic modelling in DNV and at the University of Oslo. 
In particular we want to thank Simen Eldevik, Frank Børre Pedersen and Carla Ferreira for valuable input to this paper.

\bibliography{references}

\newpage
\begin{appendices}
\section{Existence of the epistemic generator}
\label{app:epiestemic_gen_proof}

The purpose of this section, is to explain why Assumption \ref{assumption} (of the existence of a generating random variable for the epistemic $\sigma$-algebra) hold under some very mild assumptions.

In order to do this, we consider a \emph{standard probability space}. Roughly, this is a probability space consisting of an interval and/or a countable (or finite) number of atoms. Formally, a probability space is standard if it is isomorphic (up to P-null sets) with an interval equipped with the Lebesgue measure, a countable (or finite) set of atoms, or a disjoint union of both of these types of sets.

The following proposition says that in a standard probability space, any sub $\sigma$-algebra is generated by a random variable up to $P$-null sets.
For a proof, see e.g. Greinecker \cite{Greinecker}.

    \begin{prop}
    \label{prop: 1}
    Let $(\Omega, \mathcal{F}, P)$ be a standard probability space 
    and $\mathcal{E} \subseteq \mathcal{F}$ a sub $\sigma$-algebra.
    
    Then there exists an $\mathcal{F}$-measurable random variable $\theta$ such that 
    \begin{equation*}
        \mathcal{E} = \sigma(\theta) \text{ mod } 0.
    \end{equation*}
    \end{prop}
    
    Hence, as long as our probability space is standard (which is a mild assumption), we can assume that our sub $\sigma$-algebra of epistemic information, $\mathcal{E}$, is generated (up to $P$-null sets) by a random variable $\theta$ without loss of generality. Note that for the purpose of this paper, the mod 0 (i.e., up to $P$-null sets) is not a problem. Since we are only considering conditional expectations (or in particular, expectations), the $P$-null sets disappear. 
    
    Actually, this generating random variable, $\theta$, can always be modified to another random variable, $\hat{\theta}$, which is $\mathcal{E}$-measurable (purely epistemic) by augmenting the $P$-null sets. This means that $\theta$ and $\hat{\theta}$ are the same with respect to conditional expectations.
    
    Furthermore, if $\X$ is a random variable on this standard probability space, $\X | \theta$, is purely aleatory, i.e., independent of $\mathcal{E}$. This follows because $\X | \theta$ is independent of $\sigma(\theta)$ and independence holds $P$-almost surely, so mod 0 does not affect this.

\end{appendices}

\end{document}